\PassOptionsToPackage{hyphens}{url}
\documentclass[10pt]{fairmeta}

\usepackage{amsmath,amsfonts,amssymb}
\usepackage{array,tabularx,threeparttable,colortbl}
\usepackage{graphicx}
\usepackage{wrapfig}
\usepackage{lmodern}
\microtypesetup{expansion=false}
\setcitestyle{numbers,square,comma,sort&compress}
\hypersetup{
  pdftitle={4DVLT: Dynamic Scene Understanding with Worldline-Centered Vision-Language Tracking},
  pdfauthor={Boxue Yang, Chaoyue Li, Shengyao Zhou, Rui Qian, Haoyang Wu, Shaoqiu Zhang, Linfeng Zhang},
  hidelinks
}
\renewcommand{\institutionallogos}{}
\definecolor{TableNavy}{HTML}{1D4966}
\definecolor{TableHeaderBg}{HTML}{E7F0F6}
\definecolor{TableGroupBg}{HTML}{D9EAF3}
\definecolor{TableGreen}{HTML}{0B6B3D}
\definecolor{TableGreenBg}{HTML}{E5F3EA}
\definecolor{TableNeutralBg}{HTML}{F1F3F5}
\definecolor{TableMuted}{HTML}{596977}
\newcommand{\TableFont}{\small}
\newcommand{\MetricHeader}[3][]{%
  \mbox{$\mathbf{#2}\ifstrempty{#1}{}{_{\mathbf{#1}}}\!#3$}}

\newcommand{\nowrap}[1]{\mbox{#1}}
\newcommand{\deltaicon}[1]{\raisebox{0.08ex}{\ensuremath{#1}}}
\newcommand{\plaindelta}[2]{\nobreak\hspace{0.15em}\mbox{\scriptsize(#1#2)}}
\newcommand{\reportdelta}[2]{\plaindelta{#1}{#2}}

\newcommand{\updelta}[1]{\textcolor{TableGreen}{\reportdelta{\deltaicon{\uparrow}}{#1}}}

\newcolumntype{P}[1]{>{\raggedright\arraybackslash}p{#1}}
\newcolumntype{L}{>{\raggedright\arraybackslash}X}
\newcolumntype{C}{>{\centering\arraybackslash}X}
\newcolumntype{Y}[1]{>{\hsize=#1\hsize\linewidth=\hsize\centering\arraybackslash}X}

\title{4DVLT: Dynamic Scene Understanding with Worldline-Centered Vision-Language Tracking}
\renewcommand{\authorlist}{%
  \authorformat[1,*]{Boxue Yang},
  \authorformat[2,*]{Chaoyue Li},
  \authorformat[3,*]{Shengyao Zhou}\\[0.1em]
  \authorformat[4]{Rui Qian},
  \authorformat[2]{Haoyang Wu},
  \authorformat[1]{Shaoqiu Zhang},
  \authorformat[1,\dagger]{Linfeng Zhang}%
}
\renewcommand{\affiliationlist}{%
  \affiliationformat[1]{Shanghai Jiao Tong University},
  \affiliationformat[2]{Huazhong University of Science and Technology}\\
  \affiliationformat[3]{Zhejiang University},
  \affiliationformat[4]{Fudan University}%
}
\renewcommand{\contributionlist}{%
  \contributionformat[*]{Equal contribution.}\hspace{0.7em}
  \contributionformat[\dagger]{Corresponding author.}%
}
\metadata[Contact]{\email{zhanglinfeng@sjtu.edu.cn}, \email{yangboxue@sjtu.edu.cn}, \email{hustlichaoyue@hust.edu.cn}}

\begin{document}
\raggedbottom
\abstract{Language-guided dynamic scene understanding requires identifying one physical entity across time, metric 3D space, and multiple viewpoints, a setting not captured by current multimodal reasoning or tracking paradigms. We introduce \textbf{4DVLT}, an offline calibrated multi-view task that unifies grounding with metric 3D and synchronized 2D tracking through a \emph{worldline} binding identity, motion, visibility, and observations. We construct \textbf{Instruct-4D}, with 129.4K QA pairs, 64.7K targets, 851 scenes, and 9 reasoning types, and propose \textbf{4DTrack} for query-conditioned inference over a 4D state graph. 4DTrack reaches 55.18 $\mathrm{WQS}$, exceeding the strongest adapted VLT baseline by 33.01 points while improving grounding and trajectory quality.
}
\maketitle

\section{Introduction}

\begin{figure*}[t]
    \centering
    \includegraphics[width=\textwidth]{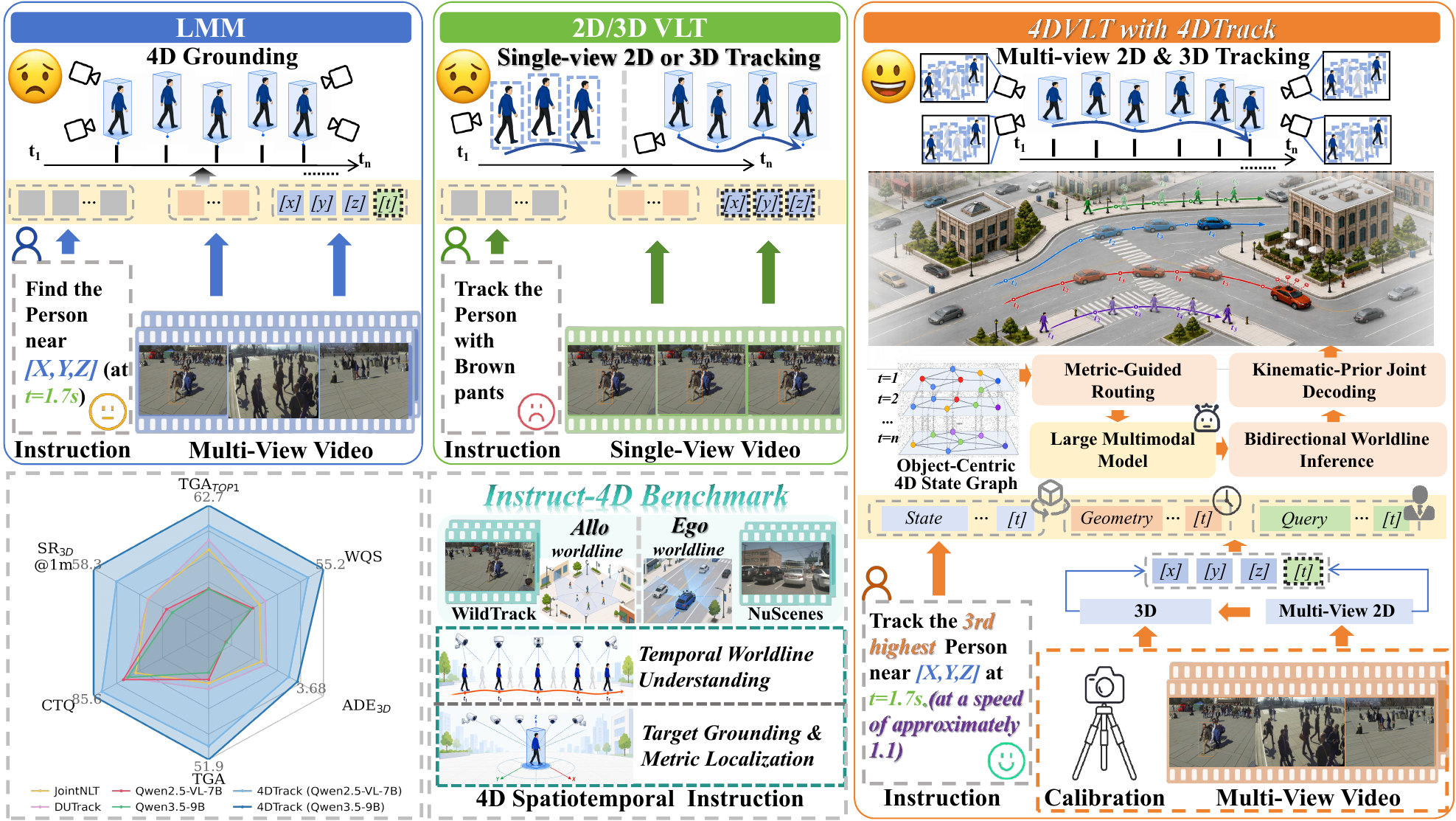}
    \caption{4DVLT unifies language grounding with metric 3D and synchronized multi-view 2D tracking. Given calibrated video and an instruction, it recovers one persistent physical entity as a worldline, binding identity, motion, visibility, and view support.}
    \label{fig:task}
\end{figure*}

Spatial intelligence requires reasoning about physical entities as they persist across time and viewpoints. Language makes these objectives inseparable: similar candidates may differ only in motion history, future behavior, or geometry across views. Large multimodal models (LMMs) exhibit broad visual understanding
\cite{liBLIP2BootstrappingLanguageimage,zhuMinigpt4EnhancingVisionlanguage,wen2026innovatorvl,li2026lmmtrack4deliciting4ddynamic},
but semantic competence does not ensure instance persistence~\cite{xu2026posreasoner}. Vision-language tracking (VLT) structures localization from language queries
\cite{liTrackingNaturalLanguage,wang2018describe,fengSiameseNaturalLanguage2021,wang2021towards,zhouJointVisualGrounding2023,maUnifyingVisualVisionlanguage2024},
yet usually separates image-plane tracking from metric 3D estimation
\cite{bertinettoFullyconvolutionalSiameseNetworks2016,zhaoEffectiveLocalGlobal2023,huangGOT10kLargeHighdiversity2021,mullerTrackingNetLargescaleDataset2018,liuMambaVLTTimeevolvingMultimodal,weiMono3DVLTMonocularvideobased3D}.
Multi-camera tracking exploits cross-view geometry but generally assumes a known target. Existing paradigms therefore leave language grounding, 2D/3D tracking, and cross-view association outside a single physical-entity inference problem.

Instructions involving \textbf{Disambiguation}, \textbf{Reverse Reasoning}, \textbf{Trajectory Shape}, and \textbf{Kinematic Shift} expose this gap: the referent may depend on nonlocal motion or a change in state. A ground-then-track pipeline prevents motion and geometry from revising an ambiguous referent, and prevents language from constraining later association. Grounding and trajectory estimation must instead be solved jointly.

Motivated by this challenge, we introduce \textbf{4DVLT} (\textbf{4D Vision-Language Tracking}), an offline task for instruction conditioned understanding and tracking in calibrated video from multiple views. 4DVLT requires a model to identify the referred physical entity, recover its metric 3D trajectory, and produce synchronized 2D tracks across views, with all outputs tied to the same persistent entity. Our key insight is that target identity, metric motion, visibility, and observations from multiple views are complementary views of one state. We formalize this state as a \textbf{worldline}, which binds the language specified identity to its 3D evolution and support across views.

To evaluate this capability, we introduce \textbf{Instruct-4D}, a benchmark with 129.4K QA pairs, 64.7K target entities, 851 scenes, and 9 query types for reasoning. Its queries cover spatiotemporal anchoring, reverse reasoning, disambiguation, 3D spatial reasoning, trajectory patterns, kinematic changes, and motion residuals. We define $\mathrm{TGA}_{\mathrm{Top1}}$, $\mathrm{TGA}$, $\mathrm{WQS}$, and $\mathrm{CTQ}$ to assess initial grounding, sequence level grounding, unconditional worldline quality, and correctly grounded trajectory quality.

To address 4DVLT, we propose \textbf{4DTrack}, a unified inference framework built on a 4D state graph over object states. Rather than treating grounding and tracking as consecutive stages, 4DTrack evaluates language, geometry, temporal order, motion continuity, and support across views over shared worldline hypotheses. Query-conditioned routing focuses the search, direction-symmetric learning captures nonlocal temporal evidence, and kinematic calibration favors coherent trajectories. On Instruct-4D, 4DTrack achieves \textbf{SOTA} performance, reaching 55.18 $\mathrm{WQS}$ and outperforming the strongest adapted VLT baseline by \textbf{33.01} points while improving target grounding and recovered 4D trajectory quality.

The contributions of this work are summarized as follows:
\begin{itemize}
    \item We introduce \textbf{4DVLT}, a unified task for understanding and tracking in dynamic multi-view scenes, where 2D and 3D predictions must remain associated with the same language-referred entity.
    \item We introduce \textbf{Instruct-4D}, a large-scale benchmark for unified instruction understanding and identity-consistent 4D tracking. 
    \item We formalize a \textbf{worldline} representation that binds target identity, metric 3D motion, visibility, and synchronized multi-view 2D observations. Based on this representation, \textbf{4DTrack} jointly performs language grounding and 2D/3D tracking over a shared 4D state graph.
    \item Experiments reveal a capability gap in existing LMM and VLT-based approaches. \textbf{4DTrack} improves target grounding and trajectory quality, outperforming the strongest adapted VLT baseline by \textbf{33.01} points in $\mathrm{WQS}$.
\end{itemize}

\begin{figure*}[t]
    \centering
    \includegraphics[width=\linewidth]{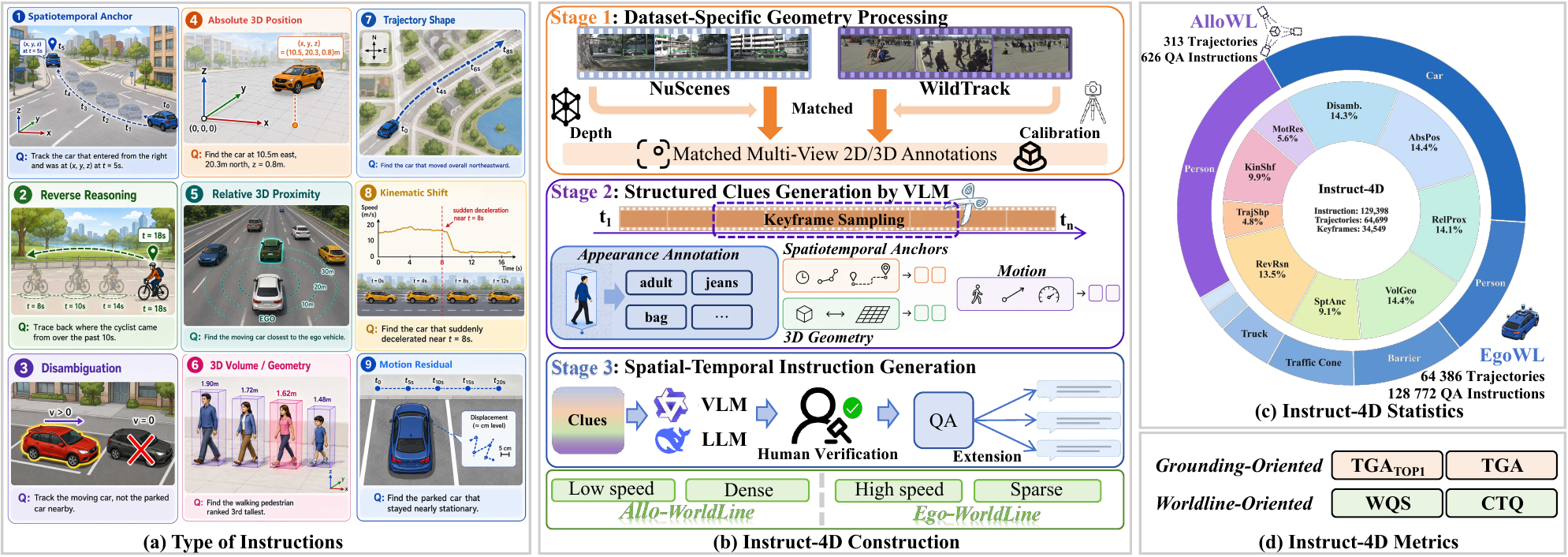}
    \caption{Instruct-4D benchmark. Instruct-4D evaluates whether a model can interpret a spatiotemporal instruction, identify the referred physical entity, and maintain identity across metric 3D space and synchronized observations from multiple views. The figure summarizes (a) nine query categories for reasoning; (b) dataset specific geometry processing, structured clue extraction, and human verified instruction generation from nuScenes and WildTrack; (c) statistics for EgoWL and AlloWL; and (d) complementary metrics for target grounding ($\mathrm{TGA}_{\mathrm{Top1}}$ and $\mathrm{TGA}$) and recovered 4D trajectory quality ($\mathrm{WQS}$ and $\mathrm{CTQ}$).}
    \label{fig:benchmark}
\end{figure*}

\section{Related Work}

\noindent \textbf{Vision-Language and Geometric Tracking.}
Vision-language tracking (VLT) associates textual descriptions with visual targets, progressing from language-guided tracking
\cite{liTrackingNaturalLanguage,fengSiameseNaturalLanguage2021}
to multimodal fusion, memory, and adaptive prompting
\cite{wang2021towards,zhouJointVisualGrounding2023,maUnifyingVisualVisionlanguage2024,zhangOnestreamVisionlanguageMemory2024,liDynamicUpdatesLanguage,liuMambaVLTTimeevolvingMultimodal}.
Recent work further explores monocular 3D VLT and language-conditioned 6-DoF point-cloud tracking
\cite{weiMono3DVLTMonocularvideobased3D,sun2024l4dtrack}
as well as LMM-assisted tracking
\cite{liDTLLMVLTDiverseText2024,zhangChatTrackerEnhancingVisual2026},
while multi-camera and 3D trackers model cross-view geometry and metric temporal states
\cite{zhang2022mutr3d,li2022time3d,feng2025st4rtrack}.
However, these settings typically separate 2D and 3D tracking or assume that the target identity is already known. In contrast, 4DVLT jointly grounds an instruction-referred entity and tracks it in metric 3D space and synchronized multi-view 2D observations.

\noindent \textbf{Large Multimodal Models for Dynamic Scene Understanding.}
Large multimodal models (LMMs) support broad visual reasoning over images and videos
\cite{liBLIP2BootstrappingLanguageimage,openaiGPT4TechnicalReport2024,zhangLLaVAvideoVideoInstruction2024,wen2025ai4service,wen2026evostreaming},
with recent 3D-aware extensions incorporating geometric and metric reasoning
\cite{zhuLLaVA3DSimpleEffective,zhengVideo3DLLMLearning,zhangFlatlandSpaceTeaching2025,zhengLearningVideos3D2025,wuMVGGTMultimodalVisual2026}.
Related work on spatiotemporal grounding and referring video segmentation predicts object tubes or masks
\cite{zhangWhereDoesIt2020,guContextguidedSpatiotemporalVideo,wuLanguageQueriesReferring2022a,wuOnlineReferSimpleOnline2023},
while emerging studies investigate 4D reasoning and dynamic objects
\cite{zhou2025llava,zhou2025vlm4d,zhou2025reason4d,yinMLLM4DVisualbasedSpatialtemporal2026,huangThinkingDynamicsHow2026,zhangUAVtrackVLAEmbodied2026}.
Nevertheless, persistent instance grounding, metric motion, and cross-view correspondence are rarely evaluated jointly. 4DVLT addresses this gap through unified instruction understanding and 4D tracking, with a worldline serving as the entity-centric representation linking identity, motion, and multi-view observations.

\section{Task Definition and Benchmark}
\label{sec:benchmark}

\subsection{4DVLT Task Definition}

\textbf{4DVLT} studies how language refers to a physical entity in a dynamic 3D scene. A query may specify the target through appearance, metric relations, temporal events, or motion patterns, and these cues can become decisive only when multiple timestamps and cameras are considered together. The prediction object is therefore the entity's \emph{worldline}, a temporally supported physical hypothesis that binds identity, 3D motion, visibility, and synchronized image observations.

Formally, given a calibrated clip from multiple views $\mathcal{V}=\{I_t^c\mid t=1,\ldots,T,\ c=1,\ldots,C\}$, camera calibration $\mathcal{K}$, and a language query $q$, a model identifies the referred target $y$ and estimates its complete worldline $\mathcal{W}_y=\{z_t\}_{t\in\mathcal{T}_y}$. Here $\mathcal{T}_y$ is the temporal support of the target, and each state $z_t$ binds semantic identity, metric 3D state, visibility, and synchronized 2D projections from multiple views. Evaluation treats the output as one physical entity that must remain consistent throughout the clip.

The complete synchronized clip is available at inference, allowing later evidence to resolve an earlier referent in Reverse Reasoning and Trajectory Shape queries. Unlike frame level 2D/3D VLT~\cite{liTrackingNaturalLanguage,zhouJointVisualGrounding2023,liuMambaVLTTimeevolvingMultimodal,weiMono3DVLTMonocularvideobased3D}, tube or mask grounding~\cite{zhangWhereDoesIt2020,guContextguidedSpatiotemporalVideo,wuLanguageQueriesReferring2022a,wuOnlineReferSimpleOnline2023}, and static 3D grounding~\cite{chenScanRefer3DObject2020}, 4DVLT evaluates whether language grounding, metric 3D reasoning, correspondence across views, and temporal coherence agree on the same recovered entity.

\subsection{Instruct-4D Benchmark}

\textbf{Instruct-4D} comprises 129.4K QA pairs over 64.7K targets from 851 scenes. It consists of two complementary subsets that capture worldline reasoning under distinct sensing configurations, as illustrated in Figure~\ref{fig:benchmark}. \textbf{Instruct-4D Ego-Worldline} (\textbf{EgoWL}), derived from nuScenes \cite{caesarNuScenesMultimodalDataset2020}, characterizes relative target motion from a moving observer. \textbf{Instruct-4D Allo-Worldline} (\textbf{AlloWL}), built upon WildTrack \cite{chavdarovaWILDTRACKMulticameraHD2018}, instead employs a fixed, world anchored camera array. Each query specifies a target identity, its complete 3D worldline, and temporally aligned 2D trajectories whenever the target is visible. EgoWL captures diverse outdoor dynamics across object categories, while AlloWL stresses pedestrian tracking in densely populated scenes with severe occlusions.

Queries are produced from calibrated tracks by converting appearance, geometry, temporal events, and motion into structured clues. Each canonical instruction is checked against the underlying geometry and retained only when it identifies one worldline; paraphrases preserve the same identity and temporal anchors. Appendix~\ref{app:data_protocol} gives the complete generation protocol.

The 129.4K corpus statistic covers all verified QA pairs; graph experiments use the subset whose proposal windows retain complete target support and satisfy the connectivity rule. Retained worldline groups follow an approximately $85{:}15$ training/validation split. EgoWL and AlloWL contribute about $99.1\%$ and $0.9\%$ of these instances, so each subset is scored before macro averaging. EgoWL partitions are scene-disjoint, whereas AlloWL uses disjoint temporal ranges and removes boundary-crossing trajectories. A worldline and all of its paraphrases remain in one partition. Reported results use the held-out validation partition of this fixed two-way protocol.

This construction yields a controlled benchmark for physically grounded 4D reasoning. Its language exposes scene ambiguity and temporal reasoning under auditable metric supervision rather than unrestricted conversational coverage.

\subsection{Evaluation Perspectives and Query Taxonomy}

The 9 query types separate metric target grounding from temporal/worldline reasoning, allowing performance differences to be interpreted by the physical evidence each instruction uses. The taxonomy also keeps generation and evaluation aligned, since each query type specifies what evidence identifies the target and which portion of the worldline must be recovered. The four metrics mirror this division through grounding and quality pairs.

\noindent \textbf{Target Grounding and Metric Localization.} Disambiguation, Absolute 3D Position, Relative 3D Proximity, and 3D Volume Geometry test whether language can identify an entity through appearance and physically meaningful relations rather than appearance alone.
\noindent \textbf{Temporal and Worldline Understanding.} Spatiotemporal Anchor, Reverse Reasoning, Trajectory Shape, Kinematic Shift, and Motion Residual test whether an instruction can constrain the target's evolution using events, ordering, path geometry, or motion change over time.

\subsection{Metrics}
\label{sec:metrics}

We use four primary metrics. For $N$ samples, let $y_i^*$ be the target identity, $m_i$ the unique modal persistent identity along the predicted path, and $m_i^{(1)}$ the identity of its first state. A tie, an empty path, or an unmatched state is incorrect. Define $\delta(a,b)=1$ if $a=b$, and $0$ otherwise.

\noindent \textbf{Target Grounding.}
Sequence level \textbf{Target Grounding Accuracy} ($\mathrm{TGA}$) and \textbf{First timestamp Target Grounding Accuracy} ($\mathrm{TGA}_{\mathrm{Top1}}$) are
\begin{equation}
\begin{aligned}
\mathrm{TGA}&=\frac{1}{N}\sum_{i=1}^{N}\delta(m_i,y_i^*),\\
\mathrm{TGA}_{\mathrm{Top1}}&=\frac{1}{N}\sum_{i=1}^{N}\delta(m_i^{(1)},y_i^*).
\end{aligned}
\label{eq:tga_metrics}
\end{equation}
They measure whether the referred entity is recovered over the full sequence and at the first aligned state, respectively.

\noindent \textbf{Worldline Quality.}
The three components reported in Table~\ref{tab:main_results} measure complete-support 3D quality (\textbf{3D}), visible-view box alignment (\textbf{2D}), and visible support coverage (\textbf{Cov.}). Their per-sample ranges are $[0,50]$, $[0,30]$, and $[0,20]$, respectively, and missing states or visible-view boxes receive zero credit. Let $Q_i=\mathrm{3D}_i+\mathrm{2D}_i+\mathrm{Cov}_i$. The \textbf{Worldline Quality Score} ($\mathrm{WQS}$) averages this decomposition over all queries, while \textbf{Conditional Tracking Quality} ($\mathrm{CTQ}$) applies the same score to correctly grounded identities:
\begin{equation}
\mathrm{WQS}=\frac{1}{N}\sum_i Q_i,\qquad
\mathrm{CTQ}=\frac{\sum_i\delta(m_i,y_i^*)Q_i}{\sum_i\delta(m_i,y_i^*)+\epsilon}.
\label{eq:quality_metrics}
\end{equation}
We also report $\mathrm{ADE}_{\mathrm{3D}}$ over valid predictions and $\mathrm{SR}_{\mathrm{3D}}@1\mathrm{m}$ over the complete target support. Appendix~\ref{app:metric_details} gives the component-level evaluator form and boundary cases.

\section{Methodology}
\label{sec:methodology_main}

\subsection{Overview}

\begin{wrapfigure}{r}{0.50\textwidth}
    \centering
    \includegraphics[width=\linewidth]{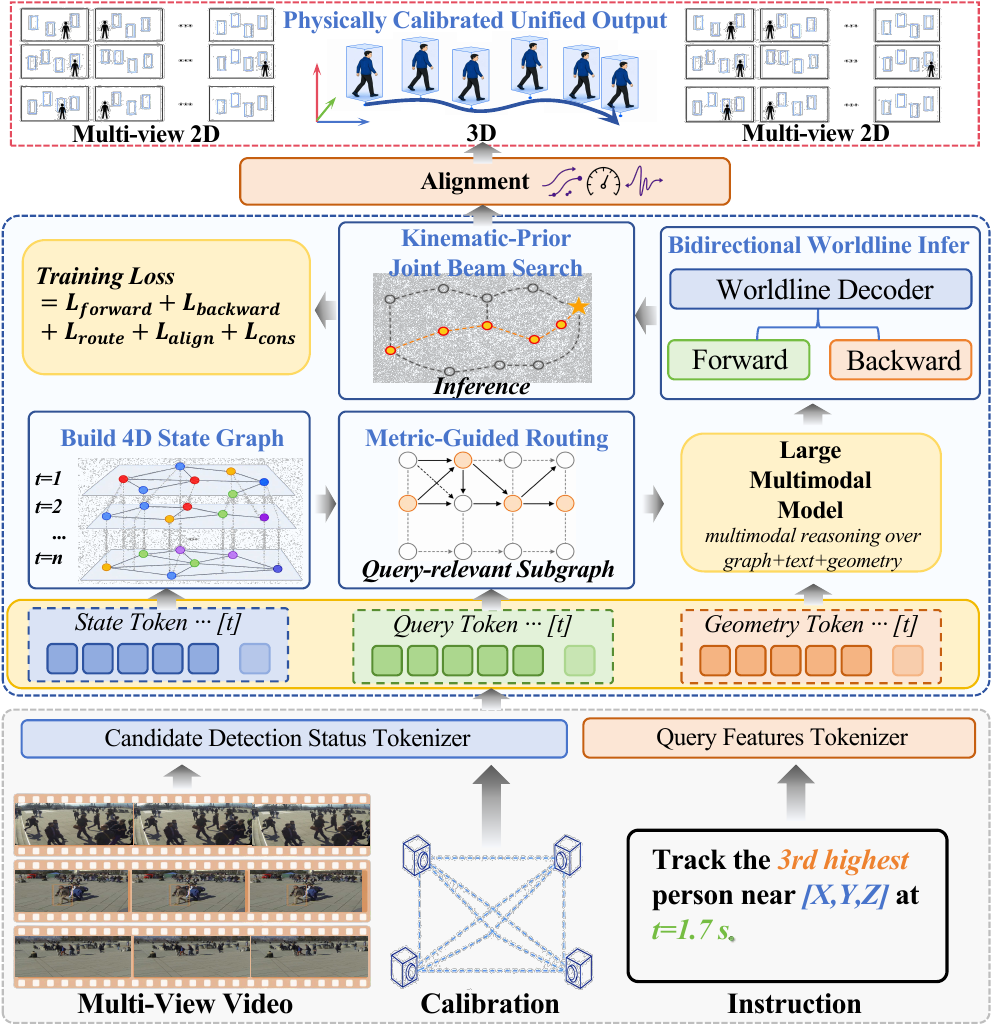}
    \caption{Overview of 4DTrack. Query conditioned routing selects a compact 4D state subgraph, and worldline decoding returns one entity trajectory with synchronized 3D and 2D outputs across views.}
    \label{fig:method}
\end{wrapfigure}

As shown in Figure~\ref{fig:method}, \textbf{4DTrack} casts \textbf{4DVLT} as query-conditioned worldline inference. The central challenge is to jointly resolve target identity from linguistic, geometric, temporal, and multi-view evidence while recovering metrically consistent 3D motion and synchronized 2D projections.

4DTrack organizes candidate object states into an object-centric 4D graph that couples metric 3D attributes with multi-view 2D evidence. Query-conditioned routing narrows this hypothesis space, direction-symmetric learning regularizes temporal recovery, and kinematic calibration selects a physically coherent path. The resulting worldline directly yields metric 3D motion and synchronized multi-view 2D tracks. Prompt construction and optimization are detailed in Appendix~\ref{app:method_details}.
\subsection{Worldline 4D State Space}

Because 4DVLT couples identity, geometry, visibility, and support across views over time, each clip is represented as a 4D graph over object states, $\mathcal{G}=(\mathcal{V}_s, \mathcal{E}_s)$. The graph defines the hypothesis space in which an entity referred to by the instruction can be recovered. Nodes are candidate object states, and temporal edges connect candidates in adjacent retained keyframes when their 3D centers satisfy a reachability radius. Selecting a target identity is thus selecting a path whose states jointly explain the 3D trajectory and visible view 2D evidence.

At timestamp $t$, the shared, query-independent front end provides $N_t$ proposal states $\{z_t^i\}_{i=1}^{N_t}$. EgoWL uses Sparse4D states in the global frame with detector-supported 2D observations; AlloWL uses the calibrated WildTrack state set. Proposals are formed before target matching and shared by every method. Each stores appearance, 3D position and extent, time, visibility, and retained 2D support, but no target flag, reference identity, or target edge enters the model. The benchmark therefore evaluates worldline inference conditional on common proposal support, not end-to-end detection. Appendix~\ref{app:graph_protocol} specifies construction and retention.

The graph preserves two forms of structure that a framewise representation discards. Node attributes bind a metric state to the camera observations of the same physical hypothesis, while temporal edges encode feasible displacement between adjacent timestamps. 4DTrack serializes frame, node, and edge blocks in temporal order and projects each node into the language-model hidden space. Explicit candidate boundaries are retained throughout: language supplies the selection criterion, but every decoded state remains tied to an auditable graph node and its synchronized observations.
\subsection{Query Conditioned Candidate Contraction}

Many \textbf{Instruct-4D} queries hinge on metric relations, interaction context, or temporal conditions rather than appearance alone. Decoding over the full graph preserves many possible paths that are irrelevant to the query, so 4DTrack contracts $\mathcal{G}$ into a relevant subgraph $\hat{\mathcal{G}}$ before worldline decoding. The contraction combines language relevance with metric anchors. Language identifies which evidence should matter, while metric anchors keep surviving candidates compatible with the physical relations stated by the query.

Given query $q$ and candidate embeddings $\{\mathbf{h}_i\}$, the router uses learned latent queries $u$. Text--node similarity gives $a_i^{\mathrm{text}}$, latent attention gives $r_{u,i}$, and a text-conditioned anchor head predicts a centered 3D anchor and clipped log-scale $\log\hat\sigma_u\in[-2,4]$. Let $\delta_{u,i}$ be the offset from this anchor. The routing logit is
\begin{equation}
\ell_{u,i} = r_{u,i} + \alpha_s a_i^{\mathrm{text}} + \alpha_m\!\left(-\frac{\|\delta_{u,i}\|_2^2}{\hat{\sigma}_u^2+\epsilon} + \beta\chi_{\mathrm{reach}}(i)\right),
\label{eq:routing_score}
\end{equation}
where $\chi_{\mathrm{reach}}(i)$ indicates that node $i$ participates in a radius-valid adjacent-frame edge and $\beta=\operatorname{softplus}(\rho)$. Latent probabilities are averaged, and the top $M=16$ states per frame form the decoding support $\hat{\mathcal G}$. Route supervision is used only during training; inference uses the predicted mask.

The latent anchors allow a query to preserve several spatially separated clues rather than collapse relevance to one text--node match. Per-frame selection maintains temporal coverage, while the reachability bias favors states supported by adjacent-frame connectivity. Contraction therefore removes unrelated branches while preserving candidates that admit coherent continuation.

\subsection{Graph-Conditioned Worldline Decoding}

Routing defines a compact hypothesis space but does not determine how its states form one identity-consistent path. The instruction and candidate graph are encoded once; the routed mask then restricts decoding to a compact support at each timestamp. At step $t$, the answer state $\mathbf h_t$ is compared with projected node embeddings $\mathbf n_i$ through
\begin{equation}
\operatorname{logit}_{t,i}=\frac{\exp(\tau)}{\sqrt D}
\left\langle \mathrm{LN}_q(W_h\mathbf h_t),\mathrm{LN}_k(\mathbf n_i)\right\rangle+M_{t,i},
\label{eq:node_decoding}
\end{equation}
where $M_{t,i}$ masks states outside the routed candidates at timestamp $t$. Time-indexed answer states share the complete graph--query context, while beam search carries accumulated semantic and kinematic evidence across the recovered path. Prediction remains on the shared proposal support: selecting a node simultaneously specifies its metric 3D state and the multi-view evidence passed to the alignment head.

\subsection{Offline Direction-Symmetric Worldline Learning}

After routing, the remaining problem is sequence level. The referred target must be recovered as one worldline, and the evidence that identifies it may appear before, after, or between the timestamps being predicted. The offline protocol makes the complete synchronized clip available, which is essential for Reverse Reasoning, Trajectory Shape, and other queries whose language describes a temporal pattern rather than a single appearance cue.

Let $\mathcal{W}=(z_1,\dots,z_L)$ and $\overleftarrow{\mathcal W}$ denote the two orderings of a target path. Suppressing shared conditioning on $q$ and $\hat{\mathcal G}$, the same decoder is supervised under both orderings as
\begin{equation}
p_d(\mathcal{W}^{(d)}) = \prod_{t=1}^{L} p_d(z_t^{(d)}\mid z_{<t}^{(d)}),\quad d\in\{f,b\}.
\label{eq:worldline_bidir}
\end{equation}
After aligning reverse distributions to absolute time, training minimizes
\begin{equation}
\mathcal L_{\mathrm{dir}}=\lambda_f\mathcal L_f+\lambda_b\mathcal L_b+
\frac{\lambda_c}{L}\sum_t\!\left[\mathrm{KL}(p_f^t\|\tilde p_b^t)+\mathrm{KL}(\tilde p_b^t\|p_f^t)\right].
\label{eq:direction_loss}
\end{equation}
The reverse ordering is a training view of the shared decoder. At inference, the complete clip and routed support condition one chronological decode, so no second trajectory or direction-reconciliation rule is introduced. The retained support fixes the output horizon, and missing states receive zero quality credit. Appendix~\ref{app:method_details} gives the complete objective.

\subsection{Physically Calibrated Unified Output}

The decoded path must remain physically plausible and consistent across views after language selection. This final step calibrates the semantic path score with a motion prior over the same graph, so the selected entity is also coherent as a metric trajectory. Let $\hat{\mathcal{P}}=\mathcal{P}(\hat{\mathcal{G}})$ be the feasible path set, and let $s_t^{m}$ and $s_t^{p}$ denote the model and kinematic step scores. We then decode
\begin{equation}
\hat{\mathcal{W}} = \arg\max_{\mathcal{W}\in\hat{\mathcal{P}}} \sum_t \big[0.70\,\tilde s_t^{m} + 0.30\,\tilde s_t^{p}\big],
\label{eq:joint_decoding}
\end{equation}
where $s_t^{m}=\log p_{\mathrm{model}}(z_{t+1}\mid t,q,\mathcal{G})$ and $s_t^{p}=-E_t(z_{t+1})$, and tildes denote normalization over the current candidates. The global coefficient is fixed for every instruction; neither query-family metadata nor an inferred category is used. The Kin. ablation in Table~\ref{tab:ablation_refine} removes this calibrated term.

Once the path is selected, the unified output is
\begin{equation}
\hat{\mathcal{Y}} = \big(\{\mathbf{c}_t\}_{t=1}^{L},\{\hat{b}_t^c\}_{t,c}\big),
\label{eq:unified_output}
\end{equation}
where $\mathbf{c}_t=\mathbf{c}(\hat{z}_t)$ denotes the 3D center of the selected state and $\hat{b}_t^c$ its view aware 2D box in camera $c$. The same selected worldline determines both outputs. The 3D trajectory is read from the selected states, while the alignment head refines retained 2D support or predicts a camera aware fallback box when support is unavailable. This coupling preserves the task contract, since grounding and synchronized 2D/3D quality are scored on the same predicted entity.

\section{Experiments}

\subsection{Experimental Setup}

\noindent \textbf{Protocol.}
\textbf{4DTrack} follows the EgoWL-to-AlloWL training schedule, and the final sequential checkpoint is evaluated on the held-out validation partition of each subset. All methods share retained timestamps, calibration, the proposal pool, and the evaluator. The default model uses Qwen3.5-9B~\cite{qwen3.5} with QLoRA~\cite{huLoRALowrankAdaptation2021,dettmersQLoRAEfficientFinetuning2023} and trainable worldline modules. Results are computed within EgoWL and AlloWL and then macro-averaged. Appendix~\ref{app:graph_protocol} gives the settings.

\subsection{Main Comparison}

\begin{table}[!htbp]
\centering
\TableFont
\def\resdeltacolor#1#2\relax{%
  \ifx#1+\textcolor{TableGreen}{(#1#2)}%
  \else\textcolor{TableMuted}{(#1#2)}\fi}
\newcommand{\resdmain}[2]{\begin{tabular}[c]{@{}c@{}}#1\\[-0.25ex]{\scriptsize\resdeltacolor#2\relax}\end{tabular}}
\begin{threeparttable}
{\setlength{\tabcolsep}{0.95pt}
\setlength{\aboverulesep}{0.20ex}
\setlength{\belowrulesep}{0.20ex}
\renewcommand{\arraystretch}{1.04}
\begin{tabularx}{\textwidth}{@{}P{0.23\textwidth}Y{1.4}*{6}{Y{0.93}}Y{1.16}*{2}{Y{0.93}}@{}}
\toprule[1.2pt]
\rowcolor{TableHeaderBg}
 &
\multicolumn{2}{c}{\textbf{Grounding}} &
\multicolumn{5}{c}{\textbf{Worldline Quality}} &
\multicolumn{3}{c}{\textbf{Tracking}} \\
\cmidrule(lr){2-3}\cmidrule(lr){4-8}\cmidrule(l){9-11}
\rowcolor{TableHeaderBg}
\multirow{-2}{*}{\textbf{Models}} & \MetricHeader[Top1]{TGA}{\uparrow} & \MetricHeader{TGA}{\uparrow}
& \MetricHeader{WQS}{\uparrow} & \MetricHeader{CTQ}{\uparrow} & \MetricHeader{3D}{\uparrow} & \MetricHeader{2D}{\uparrow} & \MetricHeader{Cov.}{\uparrow}
& \MetricHeader[3D]{ADE}{\downarrow} & \MetricHeader{SR}{\uparrow} & \MetricHeader{IDF1}{\uparrow} \\
\midrule
\rowcolor{TableGroupBg}
\multicolumn{11}{@{}c@{}}{\textcolor{TableNavy}{\textbf{Adapted VLT Trackers}}} \\
\midrule
JointNLT &37.04 &15.17 &18.89 &48.79 &~6.67 &1.37 &10.85 &~8.82 &26.08 &28.13 \\
UVLTrack &43.06 &17.43 &21.84 &55.76 &~7.41 &3.58 &10.85 &~8.33 &25.58 &29.83 \\
GLAD &42.18 &19.65 &22.08 &52.54 &~7.62 &3.61 &10.85 &~8.20 &26.08 &30.18 \\
DUTrack &42.48 &18.03 &22.17 &56.38 &~7.69 &3.63 &10.85 &~8.05 &25.66 &30.09 \\
\midrule
\rowcolor{TableGroupBg}
\multicolumn{11}{@{}c@{}}{\textcolor{TableNavy}{\textbf{Open-Source Text Models}}} \\
\midrule
Llama-3-8B &12.32 &11.93 &13.61 &59.22 &~4.90 &1.49 &~7.23 &13.60 &12.72 &16.06 \\
Mistral-7B &12.42 &10.69 &14.22 &62.99 &~4.64 &1.67 &~7.90 &13.80 &12.36 &15.73 \\
Qwen3-8B &14.56 &11.92 &14.26 &53.57 &~4.73 &1.29 &~8.24 &13.64 &12.04 &16.10 \\
\midrule
\rowcolor{TableGroupBg}
\multicolumn{11}{@{}c@{}}{\textcolor{TableNavy}{\textbf{Open-Source Multimodal Models}}} \\
\midrule
Qwen2.5-VL-7B &14.47 &13.43 &15.12 &59.70 &~5.31 &1.66 &~8.14 &13.78 &14.79 &20.06 \\
InternVL3.5-8B &17.81 &14.64 &15.99 &61.33 &~5.69 &1.77 &~8.54 &13.78 &15.24 &20.33 \\
Qwen3-VL-8B &17.60 &17.75 &17.87 &64.22 &~7.20 &2.03 &~8.65 &13.67 &18.20 &23.80 \\
Qwen3.5-9B &14.12 &10.13 &13.99 &55.90 &~4.40 &1.19 &~8.41 &13.71 &11.38 &15.78 \\
\midrule
\rowcolor{TableGroupBg}
\multicolumn{11}{@{}c@{}}{\textcolor{TableNavy}{\textbf{4DTrack Framework with Text Models}}} \\
\midrule
4DTrack-Llama-3-8B &\resdmain{25.16}{+12.84} &\resdmain{20.66}{+8.73} &\resdmain{21.45}{+7.84} &\resdmain{64.04}{+4.82} &\resdmain{~7.19}{+2.29} &\resdmain{~4.59}{+3.10} &\resdmain{~9.67}{+2.44} &\resdmain{11.93}{-1.67} &\resdmain{21.39}{+8.67} &\resdmain{28.28}{+12.22} \\
4DTrack-Mistral-7B &\resdmain{31.53}{+19.11} &\resdmain{28.04}{+17.35} &\resdmain{30.11}{+15.89} &\resdmain{67.84}{+4.85} &\resdmain{12.30}{+7.66} &\resdmain{~5.29}{+3.62} &\resdmain{12.54}{+4.64} &\resdmain{~8.05}{-5.75} &\resdmain{29.34}{+16.98} &\resdmain{36.77}{+21.04} \\
4DTrack-Qwen3-8B &\resdmain{53.27}{+38.71} &\resdmain{54.09}{+42.17} &\resdmain{51.97}{+37.71} &\resdmain{75.64}{+22.07} &\resdmain{23.09}{+18.36} &\resdmain{14.34}{+13.05} &\resdmain{14.55}{+6.31} &\resdmain{~3.74}{-9.90} &\resdmain{53.58}{+41.54} &\resdmain{60.38}{+44.28} \\
\midrule
\rowcolor{TableGroupBg}
\multicolumn{11}{@{}c@{}}{\textcolor{TableNavy}{\textbf{4DTrack Framework with Multimodal Models}}} \\
\midrule
\nowrap{4DTrack-Qwen2.5-VL-7B} &\resdmain{50.87}{+36.40} &\resdmain{45.48}{+32.05} &\resdmain{46.44}{+31.32} &\resdmain{78.48}{+18.78} &\resdmain{20.20}{+14.89} &\resdmain{11.95}{+10.29} &\resdmain{14.29}{+6.15} &\resdmain{~4.90}{-8.88} &\resdmain{44.71}{+29.92} &\resdmain{52.24}{+32.18} \\
\nowrap{4DTrack-InternVL3.5-8B} &\resdmain{25.27}{+7.46} &\resdmain{19.56}{+4.92} &\resdmain{25.65}{+9.66} &\resdmain{68.55}{+7.22} &\resdmain{~9.24}{+3.55} &\resdmain{~5.62}{+3.85} &\resdmain{10.80}{+2.26} &\resdmain{10.10}{-3.68} &\resdmain{20.83}{+5.59} &\resdmain{26.68}{+6.35} \\
\nowrap{4DTrack-Qwen3-VL-8B} &\resdmain{41.06}{+23.46} &\resdmain{32.89}{+15.14} &\resdmain{35.42}{+17.55} &\resdmain{73.83}{+9.61} &\resdmain{13.81}{+6.61} &\resdmain{~8.68}{+6.65} &\resdmain{12.93}{+4.28} &\resdmain{~7.83}{-5.84} &\resdmain{32.43}{+14.23} &\resdmain{41.04}{+17.24} \\
\rowcolor{TableGreenBg}
\textbf{4DTrack-Qwen3.5-9B} &\resdmain{\textbf{62.68}}{+48.56} &\resdmain{\textbf{51.93}}{+41.80} &\resdmain{\textbf{55.18}}{+41.19} &\resdmain{\textbf{85.57}}{+29.67} &\resdmain{\textbf{24.71}}{+20.31} &\resdmain{\textbf{16.03}}{+14.84} &\resdmain{\textbf{14.45}}{+6.04} &\resdmain{\textbf{~3.68}}{-10.03} &\resdmain{\textbf{58.27}}{+46.89} &\resdmain{\textbf{64.31}}{+48.53} \\
\bottomrule[1.2pt]
\end{tabularx}}
\caption{\textbf{Main comparison on Instruct-4D}, macro-averaged across EgoWL and AlloWL. TGA measures grounding; WQS combines 3D trajectory, visible-view 2D alignment, and coverage, with CTQ conditioned on correct grounding. ADE$_{\mathrm{3D}}$ is in meters and lower is better; all other metrics are higher-is-better. SR is $\mathrm{SR}_{\mathrm{3D}}@1\mathrm{m}$. Parentheses show the change from each matched backbone: positive changes are green and ADE reductions are shown in gray. Bold marks the default 4DTrack-Qwen3.5-9B setting.}
\label{tab:main_results}
\end{threeparttable}
\end{table}

\noindent \textbf{Baselines and Metrics.}
We compare task-adapted VLT trackers, text and multimodal controls, and matched 4DTrack variants under one output contract. JointNLT~\cite{zhouJointVisualGrounding2023}, UVLTrack~\cite{maUnifyingVisualVisionlanguage2024}, GLAD~\cite{zhangAwareDistillationRobust}, and DUTrack~\cite{liDynamicUpdatesLanguage} retain their native encoders, language fusion, and tracking objectives; a common multi-view adapter associates their 2D tracks with proposal states. Its shared visibility rule explains the equal VLT coverage, while state and box selection remain tracker-dependent. Text controls receive the same serialized states, while multimodal controls preserve their native vision encoders. Each matched pair shares the backbone, input evidence, adaptation budget, and task exposure, changing only the worldline modules. Appendix~\ref{app:adaptation} specifies the interfaces and model list.
Table~\ref{tab:main_results} compares local tracking, open source model reasoning, and worldline inference under one output contract. 4DTrack-Qwen3.5-9B leads $\mathrm{TGA}_{\mathrm{Top1}}$, $\mathrm{WQS}$, and $\mathrm{CTQ}$ at 62.68, 55.18, and 85.57; 4DTrack-Qwen3-8B reaches the highest sequence-level $\mathrm{TGA}$ of 54.09. The default improves over the strongest adapted VLT baseline by \textbf{19.62} $\mathrm{TGA}_{\mathrm{Top1}}$ and \textbf{33.01} $\mathrm{WQS}$. With Qwen3.5-9B fixed, 4DTrack gains \textbf{48.56} $\mathrm{TGA}_{\mathrm{Top1}}$ and lowers $\mathrm{ADE}_{\mathrm{3D}}$ by \textbf{10.03}\,m; the matched text and multimodal pairs show the same overall pattern. Appendix~\ref{app:subset_results} gives subset-level results.

\subsection{Query Type and Qualitative Analysis}

\begin{table}[!htbp]
\centering
\TableFont
\begin{threeparttable}
{\renewcommand{\arraystretch}{1.08}
\begin{tabularx}{\columnwidth}{@{}P{0.34\columnwidth}@{\hspace{8pt}}C@{\hspace{8pt}}C@{\hspace{8pt}}C@{\hspace{8pt}}C@{}}
\toprule[1.2pt]
\rowcolor{TableHeaderBg}
 & \multicolumn{2}{c}{\textbf{EgoWL}} & \multicolumn{2}{c}{\textbf{AlloWL}} \\
\cmidrule(lr){2-3}\cmidrule(lr){4-5}
\rowcolor{TableHeaderBg}
\multirow{-2}{*}{\textbf{Query}} & {\MetricHeader[Top1]{TGA}{\uparrow}} & {\MetricHeader[3D]{ADE}{\downarrow}} & {\MetricHeader[Top1]{TGA}{\uparrow}} & {\MetricHeader[3D]{ADE}{\downarrow}} \\
\midrule
\rowcolor{TableGroupBg}
\multicolumn{5}{@{}l}{\textcolor{TableNavy}{\textbf{Target Grounding \& Metric Localization}}}\\
\midrule
3D Volume Geometry &74.86 &4.91 &42.86 &2.22 \\
Absolute 3D Position &76.80 &4.71 &64.29 &2.26 \\
Disambiguation &76.73 &4.74 &21.43 &3.81 \\
Relative 3D Proximity &76.35 &5.00 &46.15 &2.40 \\
\midrule
\rowcolor{TableGroupBg}
\multicolumn{5}{@{}l}{\textcolor{TableNavy}{\textbf{Temporal \& Worldline Understanding}}}\\
\midrule
Spatiotemporal Anchor &73.50 &5.24 &66.67 &3.17 \\
Reverse Reasoning &74.61 &5.10 &41.18 &2.41 \\
Trajectory Shape &79.62 &4.13 &66.67 &1.75 \\
Kinematic Shift &80.03 &4.66 &44.44 &2.65 \\
Motion Residual &77.40 &4.20 &83.33 &1.03 \\
\bottomrule[1.2pt]
\end{tabularx}}
\caption{Query-type results for the default 4DTrack-Qwen3.5-9B model on Instruct-4D. TGA$_{\mathrm{Top1}}$ is reported in percent and is higher-is-better; ADE$_{\mathrm{3D}}$ measures 3D localization error in meters and is lower-is-better. EgoWL and AlloWL are reported separately to preserve their distinct sensing conditions.}
\label{tab:query_breakdown}
\end{threeparttable}
\end{table}

\begin{figure}[!htbp]
\centering
\includegraphics[width=0.85\columnwidth]{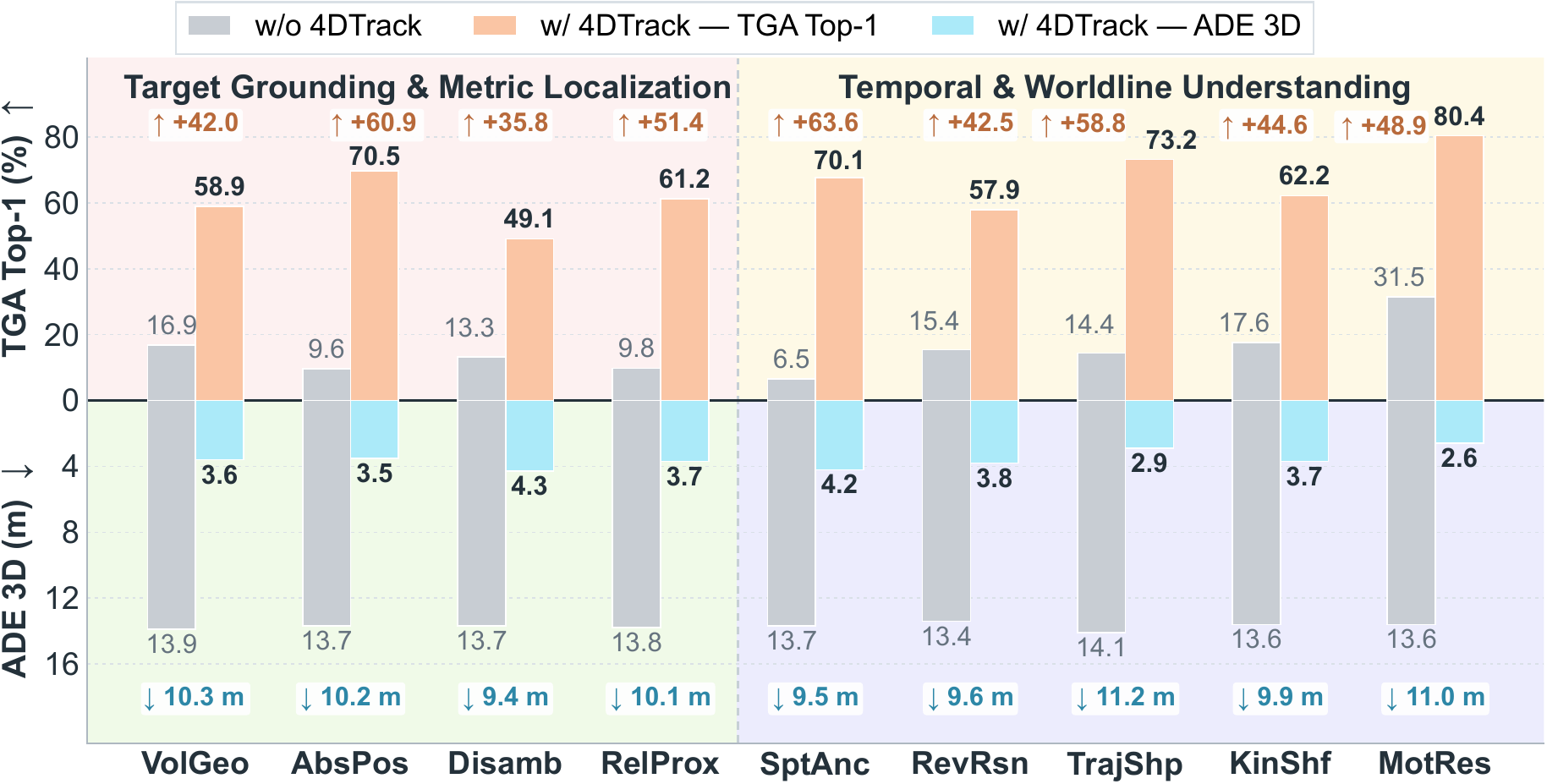}
\caption{Query level 4DTrack results, macro averaged across EgoWL and AlloWL. Labels follow Table~\ref{tab:query_breakdown}.}
\label{fig:query_breakdown}
\end{figure}

\begin{figure}[!htbp]
\centering
\includegraphics[width=0.90\linewidth]{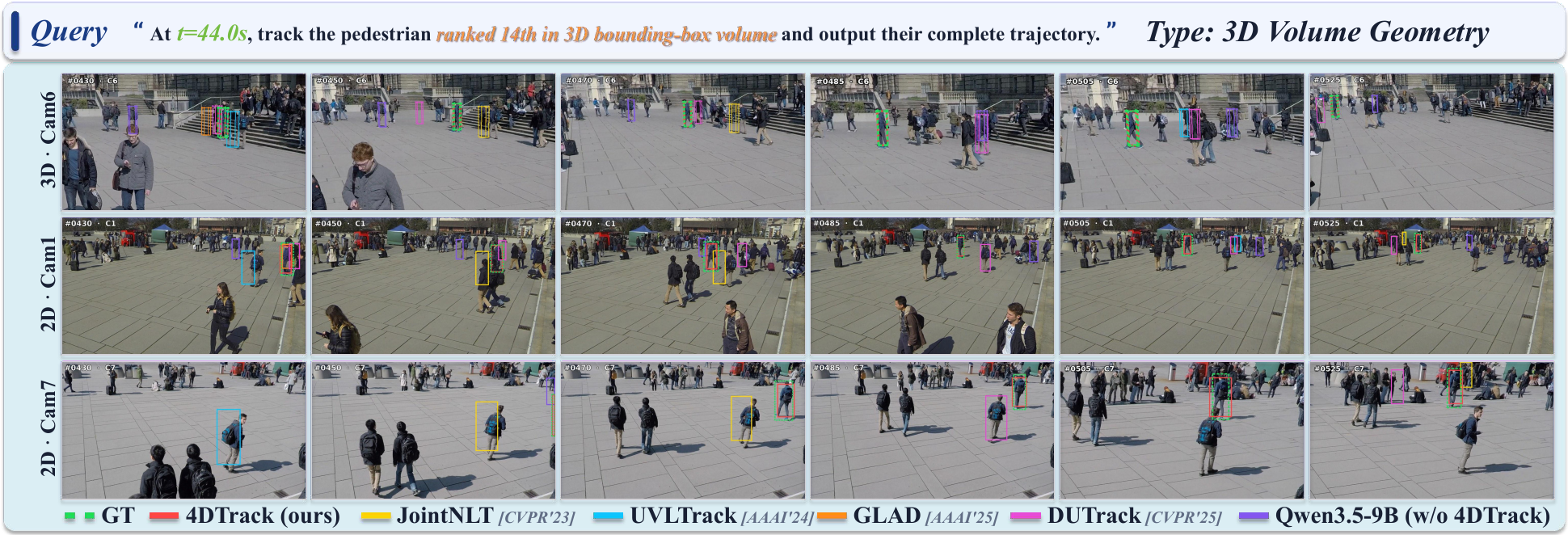}
\caption{3D Volume Geometry example from AlloWL. The query selects the pedestrian ranked 14th by 3D box volume at $t{=}44.0$\,s; columns show sampled timestamps with 3D and synchronized 2D boxes.}
\label{fig:demo}
\end{figure}

Table~\ref{tab:query_breakdown} and Figure~\ref{fig:query_breakdown} show the strongest results on motion- and metric-centered queries. Motion Residual, Trajectory Shape, and Absolute 3D Position reach macro $\mathrm{TGA}_{\mathrm{Top1}}$ of 80.37, 73.15, and 70.55, with $\mathrm{ADE}_{\mathrm{3D}}$ of 2.62, 2.94, and 3.49\,m; crowded same-category disambiguation remains the hardest AlloWL case.

Figure~\ref{fig:demo} shows an anchored 3D Volume Geometry query recovered as one synchronized worldline with 3D states and 2D tracks across views. Appendix~\ref{app:more_demos} covers the other eight query types.

\FloatBarrier
\subsection{Ablation Study for Module Analysis}

\providecommand{\abmin}[1]{#1}
\providecommand{\abpos}[2]{#1\updelta{#2}}

\newcommand{\RouteAblationTable}{
\begin{table}[!htbp]
\centering
\TableFont
\begin{threeparttable}
{\renewcommand{\arraystretch}{1.08}
\begin{tabularx}{\columnwidth}{@{}P{0.18\columnwidth}@{\hspace{6pt}}C@{\hspace{6pt}}C@{\hspace{6pt}}C@{\hspace{6pt}}C@{}}
\toprule[1.2pt]
\rowcolor{TableHeaderBg}
\textbf{Variant} 
& \MetricHeader{WQS}{\uparrow} 
& \MetricHeader{CTQ}{\uparrow} 
& \MetricHeader{TGA}{\uparrow} 
& \MetricHeader[Top1]{TGA}{\uparrow} \\
\midrule

\rowcolor{TableNeutralBg}
\abmin{Minimal} 
& \abmin{20.71} 
& \abmin{56.88} 
& \abmin{10.65} 
& \abmin{24.59} \\

Route only 
& \abpos{44.11}{23.40} 
& \abpos{69.12}{12.24} 
& \abpos{34.80}{24.15} 
& \abpos{62.43}{37.84} \\

\midrule

w/o Route 
& 29.90 
& 74.72 
& 28.38 
& 47.25 \\

\rowcolor{TableGreenBg}
\textbf{Full} 
& \abpos{\textbf{55.18}}{\textbf{25.28}} 
& \abpos{\textbf{85.57}}{\textbf{10.85}} 
& \abpos{\textbf{51.93}}{\textbf{23.55}} 
& \abpos{\textbf{62.68}}{\textbf{15.43}} \\

\bottomrule[1.2pt]
\end{tabularx}}
\caption{Routing contraction ablation on Instruct-4D. ``Route only'' adds routing to Minimal, while ``w/o Route'' removes it from Full. Green parenthesized arrows show gains over Minimal for Route only and over w/o Route for Full; all reported metrics are higher-is-better.}
\label{tab:ablation_route}
\end{threeparttable}
\end{table}
}

\newcommand{\GraphAblationTable}{
\begin{table}[!htbp]
\centering
\TableFont
\begin{threeparttable}
{\renewcommand{\arraystretch}{1.08}
\begin{tabularx}{\columnwidth}{@{}P{0.24\columnwidth}@{\hspace{2.2pt}}C@{\hspace{2.2pt}}C@{\hspace{2.2pt}}C@{\hspace{2.2pt}}C@{}}
\toprule[1.2pt]
\rowcolor{TableHeaderBg}
 & \multicolumn{2}{c}{\textbf{EgoWL}} & \multicolumn{2}{c}{\textbf{AlloWL}} \\
\cmidrule(lr){2-3}\cmidrule(lr){4-5}
\rowcolor{TableHeaderBg}
\multirow{-2}{*}{\textbf{Variant}} & \MetricHeader{WQS}{\uparrow} & \MetricHeader{CTQ}{\uparrow} & \MetricHeader{WQS}{\uparrow} & \MetricHeader{CTQ}{\uparrow} \\
\midrule
\rowcolor{TableNeutralBg}
\abmin{Minimal} & \abmin{23.02} & \abmin{74.97} & \abmin{18.40} & \abmin{38.79} \\
Graph only & \abpos{36.33}{13.31} & \abpos{80.33}{~5.36} & \abpos{20.99}{~2.59} & \abpos{51.34}{12.55} \\
\midrule
w/o Graph &52.68 &90.86 &27.50 &68.80 \\
\rowcolor{TableGreenBg}
\textbf{Full} & \abpos{\textbf{72.83}}{\textbf{20.15}} & \abpos{\textbf{93.57}}{\textbf{~2.71}} & \abpos{\textbf{37.52}}{\textbf{10.02}} & \abpos{\textbf{77.56}}{\textbf{~8.76}} \\
\bottomrule[1.2pt]
\end{tabularx}}
\caption{Object state graph ablation on Instruct-4D, shown separately for EgoWL and AlloWL. ``Graph only'' adds the graph to Minimal; ``w/o Graph'' removes it from Full. Green arrows compare each addition with its matched control. WQS and CTQ are higher-is-better.}
\label{tab:ablation_graph}
\end{threeparttable}
\end{table}
}

\newcommand{\RefineAblationTable}{
\begin{table}[!htbp]
\centering
\TableFont
\begin{threeparttable}
{\renewcommand{\arraystretch}{1.08}
\begin{tabularx}{\columnwidth}{@{}P{0.18\columnwidth}@{\hspace{6pt}}C@{\hspace{6pt}}C@{\hspace{6pt}}C@{\hspace{6pt}}C@{}}
\toprule[1.2pt]
\rowcolor{TableHeaderBg}
\textbf{Variant} & \MetricHeader[Top1]{TGA}{\uparrow} & \MetricHeader{TGA}{\uparrow} & \MetricHeader{WQS}{\uparrow} & \MetricHeader{CTQ}{\uparrow} \\
\midrule
\rowcolor{TableNeutralBg}
\abmin{Minimal} & \abmin{24.59} & \abmin{10.65} & \abmin{20.71} & \abmin{56.88} \\
Bidir. only & \abpos{36.76}{12.17} & \abpos{23.29}{12.64} & \abpos{25.24}{~4.53} & \abpos{65.93}{~9.05} \\
Kin. only & \abpos{26.80}{~2.21} & \abpos{23.15}{12.50} & \abpos{28.10}{~7.39} & \abpos{71.73}{14.85} \\
\midrule
w/o Kin. &52.09 &28.39 &38.59 &78.10 \\
\rowcolor{TableGreenBg}
\textbf{Full} & \abpos{\textbf{62.68}}{\textbf{10.59}} & \abpos{\textbf{51.93}}{\textbf{23.54}} & \abpos{\textbf{55.18}}{\textbf{16.59}} & \abpos{\textbf{85.57}}{\textbf{~7.47}} \\
\bottomrule[1.2pt]
\end{tabularx}}
\caption{Temporal and physical refinement ablation on Instruct-4D. Bidir. only and Kin. only add one component to Minimal; ``w/o Kin.'' removes kinematic scoring from Full. Green arrows show each gain over its matched control. All four metrics are higher-is-better.}
\label{tab:ablation_refine}
\end{threeparttable}
\end{table}
}

All ablations share the proposal pool, backbone, training budget, and evaluator. Minimal decodes the raw pool with forward supervision. ``Only'' rows add the named component to Minimal; ``w/o'' rows remove it from Full, which contains Graph, Route, direction-symmetric learning (Bidir.), and kinematic scoring (Kin.). Minimal and Bidir.-only differ solely in reverse-order supervision and consistency, providing the controlled without/with comparison for Bidir.

\RouteAblationTable

\noindent \textbf{Routing Contraction.}
Route only reaches 62.43 $\mathrm{TGA}_{\mathrm{Top1}}$, within 0.25 points of Full, but trails it by 17.13 $\mathrm{TGA}$, 11.07 $\mathrm{WQS}$, and 16.45 $\mathrm{CTQ}$. Routing thus resolves much of the initial ambiguity, while the remaining components recover a coherent worldline. The w/o Route row confirms its role in the full chain.

\GraphAblationTable

\noindent \textbf{Object State Graph.}
The graph over object states improves worldline quality after selection. Full exceeds w/o Graph on $\mathrm{WQS}$ and $\mathrm{CTQ}$, with larger unconditional gains on EgoWL and stronger correctly grounded gains on AlloWL.

\RefineAblationTable

\noindent \textbf{Temporal and Physical Refinement.}
Direction-symmetric learning improves nonlocal grounding over the otherwise identical Minimal model. Kin. favors physically consistent paths, and its removal lowers every primary metric after routing and graph recovery.

\FloatBarrier
\section{Conclusion}

We introduced \textbf{4DVLT}, \textbf{Instruct-4D}, and \textbf{4DTrack} for language-conditioned worldline inference in calibrated multi-view scenes. The formulation unifies semantic grounding, metric motion, and synchronized visual evidence around one persistent entity. 4DTrack separates query-independent proposals, query-conditioned contraction, direction-symmetric learning, and path calibration while preserving one auditable sequence across 3D and camera views. Separate grounding and quality scores support diagnosis under a shared output contract.

\FloatBarrier
\bibliography{reference}
\appendix

\section*{Appendix}

This appendix details the benchmark, subset results, proposal graph, implementation, evaluator, baselines, compute, and qualitative cases.

\section{Dataset Protocol}
\label{app:data_protocol}

\subsection{Partitioning}

Retained worldline groups are split approximately $85{:}15$ for training and validation. EgoWL and AlloWL contribute about $99.1\%$ and $0.9\%$ of retained graph instances. To balance these sensing regimes, main metrics are computed within each subset and then macro-averaged.

EgoWL splits disjoint scene groups from nuScenes metadata~\cite{caesarNuScenesMultimodalDataset2020}. AlloWL uses the single calibrated WildTrack capture~\cite{chavdarovaWILDTRACKMulticameraHD2018} and disjoint temporal ranges, omitting trajectories that cross a partition boundary. Instructions from one physical worldline stay in the same partition. Results use the held-out validation partition.

\subsection{Instruction Generation}

Query construction follows Figure~\ref{fig:benchmark}. First, source geometry becomes a common annotation. EgoWL uses depth, calibration, and nuScenes tracks for metric states; AlloWL uses calibrated WildTrack world-coordinate tracks. Both yield aligned 3D positions and extents, visibility, and per-view 2D boxes.

Second, informative times and intervals are selected. A vision-language model organizes appearance evidence; calibrated tracks determine metric relations, rankings, temporal anchors, trajectory shape, and motion changes. The resulting evidence spans appearance, geometry, temporal events, and motion.

Third, a VLM/LLM converts the record into an instruction. Human verification checks language, geometric agreement, and target uniqueness; ambiguous instances are removed. Each source trajectory contributes at most two canonical question--answer items. Equivalent paraphrases retain identity, query type, and temporal anchors as language augmentation, not new physical targets.

\section{Subset Results}
\label{app:subset_results}

EgoWL and AlloWL represent complementary sensing regimes rather than a common geometric scale. EgoWL contains diverse outdoor motion from a moving observer; AlloWL concentrates on crowded pedestrians, fixed cameras, and persistent occlusion. Absolute ADE should therefore be compared within a subset. Tables~\ref{tab:main_results_egowl} and~\ref{tab:main_results_allowl} expose this decomposition while preserving the main-table evaluator and model definitions. CTQ is read together with TGA because it conditions quality on correct identity, whereas WQS covers every query.

\begin{table}[!htbp]
\centering
\TableFont
\begin{threeparttable}
{\setlength{\tabcolsep}{0.95pt}
\setlength{\aboverulesep}{0.20ex}
\setlength{\belowrulesep}{0.20ex}
\renewcommand{\arraystretch}{1.04}
\begin{tabularx}{\textwidth}{@{}P{0.23\textwidth}Y{1.4}*{6}{Y{0.93}}Y{1.16}*{2}{Y{0.93}}@{}}
\toprule[1.2pt]
\rowcolor{TableHeaderBg}
 &
\multicolumn{2}{c}{\textbf{Grounding}} &
\multicolumn{5}{c}{\textbf{Worldline Quality}} &
\multicolumn{3}{c}{\textbf{Tracking}} \\
\cmidrule(lr){2-3}\cmidrule(lr){4-8}\cmidrule(l){9-11}
\rowcolor{TableHeaderBg}
\multirow{-2}{*}{\textbf{Models}} & \MetricHeader[Top1]{TGA}{\uparrow} & \MetricHeader{TGA}{\uparrow}
& \MetricHeader{WQS}{\uparrow} & \MetricHeader{CTQ}{\uparrow} & \MetricHeader{3D}{\uparrow} & \MetricHeader{2D}{\uparrow} & \MetricHeader{Cov.}{\uparrow}
& \MetricHeader[3D]{ADE}{\downarrow} & \MetricHeader{SR}{\uparrow} & \MetricHeader{IDF1}{\uparrow} \\
\midrule
\rowcolor{TableGroupBg}
\multicolumn{11}{@{}c@{}}{\textcolor{TableNavy}{\textbf{Adapted VLT Trackers}}} \\
JointNLT &40.74 &21.51 &22.34 &54.76 &8.89 &1.86 &11.59 &13.78 &28.53 &31.29 \\
UVLTrack &47.87 &24.07 &26.24 &62.12 &9.83 &4.82 &11.59 &12.68 &31.59 &33.33 \\
GLAD &48.09 &24.58 &26.37 &61.90 &9.96 &4.82 &11.59 &12.49 &31.79 &33.52 \\
DUTrack &47.71 &25.27 &26.70 &62.18 &10.26 &4.84 &11.59 &12.13 &32.27 &34.37 \\
\midrule
\rowcolor{TableGroupBg}
\multicolumn{11}{@{}c@{}}{\textcolor{TableNavy}{\textbf{Open-Source Text Models}}} \\
Llama-3-8B &12.87 &12.09 &14.23 &73.17 &5.46 &1.62 &7.15 &23.31 &12.37 &15.41 \\
Mistral-7B &12.08 &11.57 &14.50 &78.93 &5.30 &2.05 &7.15 &23.46 &11.75 &14.23 \\
Qwen3-8B &14.41 &13.06 &14.82 &72.53 &6.07 &1.62 &7.12 &23.11 &13.54 &16.87 \\
\midrule
\rowcolor{TableGroupBg}
\multicolumn{11}{@{}c@{}}{\textcolor{TableNavy}{\textbf{Open-Source Multimodal Models}}} \\
Qwen2.5-VL-7B &12.26 &11.17 &13.61 &73.95 &5.14 &1.71 &6.76 &23.92 &11.88 &14.81 \\
InternVL3.5-8B &12.08 &11.62 &14.00 &74.90 &5.25 &1.75 &7.01 &23.98 &11.97 &15.09 \\
Qwen3-VL-8B &11.66 &10.98 &13.61 &74.85 &4.96 &1.80 &6.85 &24.21 &11.40 &14.21 \\
Qwen3.5-9B &12.54 &11.43 &13.83 &72.07 &4.91 &1.64 &7.28 &23.62 &11.51 &15.02 \\
\midrule
\rowcolor{TableGroupBg}
\multicolumn{11}{@{}c@{}}{\textcolor{TableNavy}{\textbf{4DTrack Framework with Text Models}}} \\
4DTrack-Llama-3-8B &14.04 &10.93 &14.66 &71.82 &4.48 &2.50 &7.68 &21.26 &12.84 &14.46 \\
4DTrack-Mistral-7B &41.48 &35.49 &34.91 &77.05 &16.35 &6.49 &12.08 &13.11 &37.12 &42.58 \\
4DTrack-Qwen3-8B &71.24 &67.98 &69.44 &93.32 &33.65 &19.54 &16.25 &5.27 &69.20 &74.86 \\
\midrule
\rowcolor{TableGroupBg}
\multicolumn{11}{@{}c@{}}{\textcolor{TableNavy}{\textbf{4DTrack Framework with Multimodal Models}}} \\
\nowrap{4DTrack-Qwen2.5-VL-7B} &63.49 &59.58 &61.14 &92.30 &29.43 &16.64 &15.07 &7.22 &60.03 &64.86 \\
\nowrap{4DTrack-InternVL3.5-8B} &34.84 &32.25 &35.49 &90.86 &15.49 &9.16 &10.84 &15.91 &31.98 &36.38 \\
\nowrap{4DTrack-Qwen3-VL-8B} &41.91 &38.33 &41.73 &89.85 &18.52 &10.85 &12.36 &12.96 &38.10 &44.63 \\
\rowcolor{TableGreenBg}
\textbf{4DTrack-Qwen3.5-9B} &\textbf{76.34} &\textbf{69.55} &\textbf{72.83} &\textbf{93.57} &\textbf{35.59} &\textbf{20.64} &\textbf{16.60} &\textbf{4.83} &\textbf{74.62} &\textbf{77.12} \\
\bottomrule[1.2pt]
\end{tabularx}}
\caption{\textbf{EgoWL subset comparison} on the held-out validation partition. Columns and model groups follow Table~\ref{tab:main_results}. The 3D, 2D, and coverage terms sum to WQS up to rounding; ADE$_{\mathrm{3D}}$ is measured in meters and lower is better. Bold identifies the default 4DTrack-Qwen3.5-9B row.}
\label{tab:main_results_egowl}
\end{threeparttable}
\end{table}

\begin{table}[!htbp]
\centering
\TableFont
\begin{threeparttable}
{\setlength{\tabcolsep}{0.95pt}
\setlength{\aboverulesep}{0.20ex}
\setlength{\belowrulesep}{0.20ex}
\renewcommand{\arraystretch}{1.04}
\begin{tabularx}{\textwidth}{@{}P{0.23\textwidth}Y{1.4}*{6}{Y{0.93}}Y{1.16}*{2}{Y{0.93}}@{}}
\toprule[1.2pt]
\rowcolor{TableHeaderBg}
 &
\multicolumn{2}{c}{\textbf{Grounding}} &
\multicolumn{5}{c}{\textbf{Worldline Quality}} &
\multicolumn{3}{c}{\textbf{Tracking}} \\
\cmidrule(lr){2-3}\cmidrule(lr){4-8}\cmidrule(l){9-11}
\rowcolor{TableHeaderBg}
\multirow{-2}{*}{\textbf{Models}} & \MetricHeader[Top1]{TGA}{\uparrow} & \MetricHeader{TGA}{\uparrow}
& \MetricHeader{WQS}{\uparrow} & \MetricHeader{CTQ}{\uparrow} & \MetricHeader{3D}{\uparrow} & \MetricHeader{2D}{\uparrow} & \MetricHeader{Cov.}{\uparrow}
& \MetricHeader[3D]{ADE}{\downarrow} & \MetricHeader{SR}{\uparrow} & \MetricHeader{IDF1}{\uparrow} \\
\midrule
\rowcolor{TableGroupBg}
\multicolumn{11}{@{}c@{}}{\textcolor{TableNavy}{\textbf{Adapted VLT Trackers}}} \\
JointNLT &33.33 &8.82 &15.43 &42.81 &4.44 &0.87 &10.11 &3.85 &23.62 &24.98 \\
UVLTrack &38.24 &10.78 &17.43 &49.39 &4.99 &2.33 &10.11 &3.97 &19.57 &26.32 \\
GLAD &36.27 &14.71 &17.79 &43.17 &5.28 &2.40 &10.11 &3.91 &20.37 &26.83 \\
DUTrack &37.25 &10.78 &17.63 &50.58 &5.11 &2.41 &10.11 &3.96 &19.04 &25.81 \\
\midrule
\rowcolor{TableGroupBg}
\multicolumn{11}{@{}c@{}}{\textcolor{TableNavy}{\textbf{Open-Source Text Models}}} \\
Llama-3-8B &11.76 &11.76 &12.98 &45.27 &4.33 &1.35 &7.30 &3.88 &13.07 &16.72 \\
Mistral-7B &12.75 &9.80 &13.93 &47.04 &3.98 &1.29 &8.65 &4.14 &12.96 &17.23 \\
Qwen3-8B &14.71 &10.78 &13.70 &34.61 &3.38 &0.96 &9.36 &4.17 &10.53 &15.32 \\
\midrule
\rowcolor{TableGroupBg}
\multicolumn{11}{@{}c@{}}{\textcolor{TableNavy}{\textbf{Open-Source Multimodal Models}}} \\
Qwen2.5-VL-7B &16.67 &15.69 &16.62 &45.45 &5.49 &1.61 &9.52 &3.63 &17.69 &25.30 \\
InternVL3.5-8B &23.53 &17.65 &17.98 &47.76 &6.13 &1.79 &10.06 &3.57 &18.50 &25.56 \\
Qwen3-VL-8B &23.52 &24.51 &22.12 &53.59 &9.43 &2.25 &10.45 &3.13 &24.99 &33.38 \\
Qwen3.5-9B &15.69 &8.82 &14.15 &39.72 &3.88 &0.73 &9.54 &3.79 &11.25 &16.53 \\
\midrule
\rowcolor{TableGroupBg}
\multicolumn{11}{@{}c@{}}{\textcolor{TableNavy}{\textbf{4DTrack Framework with Text Models}}} \\
4DTrack-Llama-3-8B &36.27 &30.39 &28.24 &56.25 &9.90 &6.68 &11.66 &2.59 &29.94 &42.09 \\
4DTrack-Mistral-7B &21.57 &20.59 &25.31 &58.62 &8.24 &4.08 &12.99 &2.98 &21.56 &30.96 \\
4DTrack-Qwen3-8B &35.29 &\textbf{40.20} &34.49 &57.95 &12.52 &9.13 &12.84 &\textbf{2.20} &37.96 &45.90 \\
\midrule
\rowcolor{TableGroupBg}
\multicolumn{11}{@{}c@{}}{\textcolor{TableNavy}{\textbf{4DTrack Framework with Multimodal Models}}} \\
\nowrap{4DTrack-Qwen2.5-VL-7B} &38.24 &31.37 &31.74 &64.65 &10.97 &7.26 &\textbf{13.51} &2.58 &29.39 &39.61 \\
\nowrap{4DTrack-InternVL3.5-8B} &15.69 &6.86 &15.81 &46.23 &2.98 &2.08 &10.76 &4.28 &9.67 &16.97 \\
\nowrap{4DTrack-Qwen3-VL-8B} &40.20 &27.45 &29.11 &57.80 &9.10 &6.51 &13.50 &2.69 &26.75 &37.44 \\
\rowcolor{TableGreenBg}
\textbf{4DTrack-Qwen3.5-9B} &\textbf{49.02} &34.31 &\textbf{37.52} &\textbf{77.56} &\textbf{13.82} &\textbf{11.42} &12.29 &2.52 &\textbf{41.91} &\textbf{51.49} \\
\bottomrule[1.2pt]
\end{tabularx}}
\caption{\textbf{AlloWL subset comparison} on the held-out validation partition. Model groups and metric directions match the EgoWL table; ADE$_{\mathrm{3D}}$ is in meters and lower is better. The two subsets are reported separately because their sensing geometry and trajectory scales differ. Bold highlights the strongest displayed result in each metric column.}
\label{tab:main_results_allowl}
\end{threeparttable}
\end{table}

\section{Shared Candidate Graph}
\label{app:graph_protocol}

All methods operate on the same query-independent proposal graph. Each query window contains at most 20 retained frames and 24 candidate states per frame. EgoWL uses confidence-ranked Sparse4D proposals transformed to the global frame and detector-supported 2D observations. AlloWL uses the calibrated WildTrack state set as its proposal bank. Cached runs serialize these source states without consulting the query. A node stores appearance, metric position and extent, time, visibility, and per-view 2D support. Persistent identifiers and target flags are reserved for supervision and evaluation and are never embedded in model input.

After proposal construction, target annotations are matched to candidate states to define training labels. A window is retained only if every supervised state is present and consecutive supervised states are connected by the graph rule. This selection does not insert target nodes or target edges into the inference graph. At evaluation, the model receives the instruction and proposal graph but no target identifier, target flag, or reference worldline.

Adjacent-frame nodes are connected when their 3D centers are within 8\,m; each edge stores relative displacement and distance. The router uses 32 latent queries and retains the $M=16$ highest-scoring candidates at each frame. During training only, supervised nodes are merged into the decoding mask to stabilize early optimization. Validation uses the predicted mask without this merge. Beam search retains four partial paths.

\begingroup
\setlength{\abovedisplayskip}{3pt plus 1pt minus 1pt}
\setlength{\belowdisplayskip}{3pt plus 1pt minus 1pt}
\setlength{\abovedisplayshortskip}{0pt}
\setlength{\belowdisplayshortskip}{2pt plus 1pt minus 1pt}
\section{Implementation Details}
\label{app:method_details}

\subsection{Graph-Conditioned Backbone}

The candidate graph and instruction form a single causal sequence,
\begin{equation}
\mathcal{S}=\operatorname{concat}\!\left(\tau_g,\mathrm{Seq}(\mathcal{G}),
\tau_{\bar g},\tau_q,q,\tau_{\bar q},\tau_a\right),
\label{eq:prompt_serialization}
\end{equation}
where $\mathrm{Seq}(\mathcal{G})$ lists compact frame, node, and edge blocks and the $\tau$ symbols delimit graph, query, and answer regions. A discrete node token reserves its sequence position; the following soft slot receives the projected node representation,
\begin{equation}
\mathbf{H}^{(0)}[\mathrm{pos}_i]\leftarrow
\phi_{\mathrm{proj}}(\mathbf{a}_i,\mathbf{g}_i,\mathbf{u}_i),
\label{eq:soft_injection}
\end{equation}
with appearance $\mathbf a_i$, geometry $\mathbf g_i$, and auxiliary attributes $\mathbf u_i$. The projector is
\begin{equation}
\phi_{\mathrm{proj}}(\mathbf a,\mathbf g,\mathbf u)=
W_2\,\mathrm{LN}\!\left(\mathrm{GELU}(W_1[\mathbf a;\mathbf g;\mathbf u])\right).
\label{eq:graph_projector}
\end{equation}

The default language model is a 4-bit Qwen3.5-9B backbone~\cite{qwen3.5}. Low-rank adapters~\cite{huLoRALowrankAdaptation2021} update the self-attention query and value projections,
\begin{equation}
W'_{k,\ell}=W_{k,\ell}+\frac{\alpha_{\mathrm{LoRA}}}{r}B_{k,\ell}A_{k,\ell},
\qquad k\in\{q,v\}.
\label{eq:lora_update}
\end{equation}

\subsection{Metric-Guided Routing and Node Decoding}

Let $\mathbf t$ be the graph--query context state immediately before the answer region and $\mathbf n_i$ the projected node embedding, both of dimension $D$. Their global relevance is
\begin{equation}
a_i^{\mathrm{text}}=\frac{\langle W_t\mathbf t,W_n\mathbf n_i\rangle}{\sqrt D}.
\label{eq:text_node_score}
\end{equation}
For each learned latent query $\mathbf q_u$, an anchor head predicts a centered metric position and scale,
\begin{equation}
[\hat{\mathbf p}_u;\log\hat\sigma_u]=
\phi_{\mathrm{anc}}([\mathbf q_u;\mathbf t]),
\label{eq:anchor_scale}
\end{equation}
where $\log\hat\sigma_u$ is clipped to $[-2,4]$. Let $\bar{\mathbf p}_{\mathrm{valid}}$ be the mean valid position, $\delta_i=\mathbf p_i-\bar{\mathbf p}_{\mathrm{valid}}$, and $\delta_{u,i}=\delta_i-\hat{\mathbf p}_u$. The metric bias is
\begin{equation}
m_{u,i}=-\frac{\|\delta_{u,i}\|_2^2}{\hat\sigma_u^2+\epsilon}
+\beta\chi_{\mathrm{reach}}(i),\qquad \beta=\mathrm{softplus}(\rho).
\label{eq:centered_metric_bias}
\end{equation}
Here $\chi_{\mathrm{reach}}(i)=1$ when node $i$ participates in a radius-valid adjacent-frame edge. With head-averaged latent similarity $r_{u,i}$,
\begin{equation}
\begin{aligned}
\ell_{u,i}&=r_{u,i}+\alpha_s a_i^{\mathrm{text}}+\alpha_m m_{u,i},\\
w_i&=\frac1{N_q}\sum_u\operatorname{softmax}(\ell_u)_i.
\end{aligned}
\label{eq:route_logits}
\end{equation}
Per-frame top-$M$ selection defines the keep mask and the induced decoding support $\hat{\mathcal G}$.

During training, teacher-forced node tokens provide the state for worldline step $t$ at answer offset $s$,
\begin{equation}
\mathbf h_t=\mathbf H^{[\mathrm{last}]}[s-1+t].
\label{eq:answer_hidden}
\end{equation}
Let $\bar{\mathbf q}_t=\mathrm{LN}_q(W_h\mathbf h_t)$, $\bar{\mathbf k}_i=\mathrm{LN}_k(\mathbf n_i)$, and $\tau'=\exp(\tau)$. The node score is
\begin{equation}
\operatorname{logit}_{t,i}=\frac{\tau'}{\sqrt D}
\langle\bar{\mathbf q}_t,\bar{\mathbf k}_i\rangle+M_{t,i},
\label{eq:node_head}
\end{equation}
At inference, the graph--query context state is combined with a learned time embedding to produce the corresponding $\mathbf h_t$. In both cases, $M_{t,i}=0$ for routed nodes at frame $t$ and $-10^4$ otherwise.

\subsection{View-Aware Alignment}

For state $t$ and camera $c$, let $\psi_t^c$ be its retained support box and $\mathbf 1_t^c$ its validity. With camera embedding $\mathbf e_{\mathrm{cam}}(c)$ and $\mathbf x_t^c=[\mathbf h_t;\mathbf e_{\mathrm{cam}}(c);\psi_t^c;\mathbf 1_t^c]$, the refinement is
\begin{equation}
\Delta b_t^c=\Delta_{\max}\tanh\!\left(\phi_{\mathrm{res}}(\mathbf x_t^c)\right).
\label{eq:align_residual}
\end{equation}
Let $\bar b_t^c=\sigma(\phi_{\mathrm{fb}}([\mathbf h_t;\mathbf e_{\mathrm{cam}}(c)]))$ and $\tilde b_t^c=\operatorname{clip}_{[0,1]^4}(\psi_t^c+\Delta b_t^c)$. The output is
\begin{equation}
\hat b_t^c=\bar b_t^c+\mathbf 1_t^c(\tilde b_t^c-\bar b_t^c).
\label{eq:align_box}
\end{equation}
Visible support is refined locally; otherwise the head uses a camera-conditioned fallback. At inference, both $\psi_t^c$ and $\mathbf 1_t^c$ come from the selected proposal node.

\subsection{Direction-Symmetric Objective}

Let $\mathcal W^*$ be the target worldline and $\overleftarrow{\mathcal W}^*$ its reverse ordering. These are two training views of one shared decoder. Teacher-forced forward and reverse losses are
\begin{align}
\mathcal L_f&=-\sum_{t=1}^{L}\log p_f(z_t^*\mid z_{<t}^*),\label{eq:forward_loss}\\
\mathcal L_b&=-\sum_{t=1}^{L}\log p_b(\bar z_t^*\mid\bar z_{<t}^*).\label{eq:backward_loss}
\end{align}
Let $\tilde p_b^t=\Pi(p_b^{L-t+1})$ align reverse predictions to the corresponding absolute time, and $d(p,q)=\mathrm{KL}(p\|q)+\mathrm{KL}(q\|p)$. Cross-direction consistency is
\begin{equation}
\mathcal L_{\mathrm{cons}}=\frac1{L}\sum_{t=1}^{L}d(p_f^t,\tilde p_b^t).
\label{eq:consistency_loss}
\end{equation}
The reverse ordering is used only to shape the shared decoder during training. At inference, the decoder produces chronological logits from the complete graph--query context; beam search introduces path dependence through accumulated model and kinematic scores and returns one forward-time path. Thus no second trajectory or post-hoc direction reconciliation is required.

The router is supervised with binary cross-entropy,
\begin{equation}
\mathcal L_{\mathrm{route}}=\frac1{BN}\sum_{b,i}
\operatorname{BCE}(\tilde y_{b,i},\tilde w_{b,i}),
\label{eq:route_loss}
\end{equation}
where $\tilde y_{b,i}=1/L_b$ for a target-worldline node and zero otherwise, and $\tilde w$ is clipped to $[\epsilon,1-\epsilon]$. With $m_t^c$ denoting visible-view supervision and $\ell_{2D}=\lambda_1\mathrm{SmoothL1}+\lambda_{\mathrm{iou}}(1-\mathrm{IoU})$,
\begin{equation}
\mathcal L_{\mathrm{align}}=
\frac{\sum_{t,c}m_t^c\ell_{2D}(\hat b_t^c,b_t^{c,*})}{\sum_{t,c}m_t^c}.
\label{eq:align_loss}
\end{equation}
The full objective is
\begin{equation}
\mathcal L=\lambda_f\mathcal L_f+\lambda_b\mathcal L_b+
\lambda_r\mathcal L_{\mathrm{route}}+\lambda_a\mathcal L_{\mathrm{align}}+
\lambda_c\mathcal L_{\mathrm{cons}}.
\label{eq:total_loss}
\end{equation}

\subsection{Kinematic Beam Search}

For successor $j$ after $z_{t-1},z_t$, define
\begin{equation}
\mathbf v_t(j)=\mathbf p_j-\mathbf p_{z_t},\qquad
\mathbf a_t(j)=\mathbf v_t(j)-(\mathbf p_{z_t}-\mathbf p_{z_{t-1}}),
\label{eq:kin_av}
\end{equation}
and
\begin{equation}
E_t(j)=\tau_v\|\mathbf v_t(j)\|_2+\tau_a\|\mathbf a_t(j)\|_2.
\label{eq:kin_energy}
\end{equation}
Both subsets use metric coordinates on a synchronized 2\,Hz grid. We set $\tau_v=0.5$ and $\tau_a=1.0$, giving $\log p_{\mathrm{phys}}=-E_t$. With model and physics log scores renormalized over the current routed candidates, beam search accumulates
\begin{equation}
\operatorname{score}(z_{0:t})=\sum_{\tau\le t}
\left[0.70\,\tilde\ell_\tau^m+0.30\,\tilde\ell_\tau^p\right].
\label{eq:beam_score}
\end{equation}
The coefficient is fixed globally for every instruction. Decoding receives neither the benchmark query type nor an inferred family label.

\subsection{Optimization}

QLoRA~\cite{dettmersQLoRAEfficientFinetuning2023} uses rank 64, scaling 128, dropout 0.05, and query/value projection targets. Computation uses bf16, with learning rates $2{\times}10^{-4}$ for LoRA and $10^{-3}$ for the projector, router, node head, and alignment head. Training proceeds in three stages. Stage 1 freezes LoRA and warms up routing with $\lambda_f=0.1$, $\lambda_b=\lambda_c=0$. Stage 2 enables LoRA and matched forward/reverse supervision with $\lambda_f=\lambda_b=1$ and $\lambda_c=0$. Stage 3 activates consistency with $\lambda_c=0.2$. Early sampling emphasizes anchored queries and short paths; later sampling increases temporally distributed queries and difficult cases with close neighbors, visibility changes, or large second-order motion.

\paragraph{Ablation correspondence.}
Minimal and Bidir.-only form the controlled without/with comparison for direction-symmetric learning: both use the same raw pool, backbone, optimization budget, forward objective, and evaluator; Bidir.-only adds $\mathcal L_b$ and $\mathcal L_{\mathrm{cons}}$. Both return one chronological path. Graph, Route, and Kin. are analyzed through the corresponding only or leave-one-out configurations.

\section{Evaluator Details}
\label{app:metric_details}

Let $\mathcal T_i^*$ be the complete target support and $\Omega_i\subseteq\mathcal T_i^*$ the timestamps with a valid 3D prediction. Set $\hat r_{i,t}=\mathbf 1[t\in\Omega_i]$. The diagnostic displacement error is
\begin{equation}
\mathrm{ADE}_{3D,i}=\frac1{|\Omega_i|}\sum_{t\in\Omega_i}
\|\mathbf c_{i,t}-\mathbf c_{i,t}^*\|_2,
\label{eq:ade3d}
\end{equation}
and is undefined for an empty prediction. The bounded 3D quality term instead uses the complete support,
\begin{equation}
q_i^{3D}=\frac1{|\mathcal T_i^*|}\sum_{t\in\mathcal T_i^*}
\hat r_{i,t}\exp\!\left(-\frac{\|\mathbf c_{i,t}-\mathbf c_{i,t}^*\|_2}{1\,\mathrm m}\right).
\label{eq:q3d}
\end{equation}
Thus a missing state contributes zero rather than disappearing under prefix alignment.

For ground-truth visibility $v_{i,t}^c$, prediction validity $\hat v_{i,t}^c$, predicted box $\hat b_{i,t}^c$, and target box $b_{i,t}^{c,*}$,
\begin{align}
I_i&=\frac{\sum_{t,c}v_{i,t}^c\hat v_{i,t}^c
\operatorname{IoU}(\hat b_{i,t}^c,b_{i,t}^{c,*})}{\sum_{t,c}v_{i,t}^c},\nonumber\\
\mathrm{VCov}_i&=\frac{\sum_{t,c}v_{i,t}^c\hat v_{i,t}^c}{\sum_{t,c}v_{i,t}^c}.
\label{eq:visible_components}
\end{align}
The three weighted contributions reported in the main table are
\begin{equation}
\mathrm{3D}_i=50q_i^{3D},\qquad
\mathrm{2D}_i=30I_i,\qquad
\mathrm{Cov}_i=20\mathrm{VCov}_i,
\label{eq:quality_components}
\end{equation}
and the per-sample score is
\begin{equation}
Q_i=\mathrm{3D}_i+\mathrm{2D}_i+\mathrm{Cov}_i.
\label{eq:sample_quality}
\end{equation}
These are precisely the \textbf{3D}, \textbf{2D}, and \textbf{Cov.} columns in Table~\ref{tab:main_results}; each row therefore satisfies $\mathrm{WQS}=\mathrm{3D}+\mathrm{2D}+\mathrm{Cov.}$ up to rounding. If no target view is visible, the 2D and Cov. terms are set to zero. Finally,
\begin{equation}
\mathrm{SR}_{3D}@1\mathrm m=\frac1N\sum_i\frac1{|\mathcal T_i^*|}
\sum_{t\in\mathcal T_i^*}\hat r_{i,t}
\mathbf 1[\|\mathbf c_{i,t}-\mathbf c_{i,t}^*\|_2\le1\,\mathrm m].
\label{eq:sr3d}
\end{equation}
SR is therefore a temporal success rate over full target support, whereas ADE summarizes valid predictions. IDF1 is the identity F1 over timestamp-aligned states under a 2\,m 3D association gate; it is diagnostic and does not enter WQS. Under the reported evaluator protocol, sequence-level TGA is one sample-level decision: the unique modal persistent identifier along the predicted path must equal the target; a tie or missing identity is incorrect. TGA$_{\mathrm{Top1}}$ uses the first predicted node. Identity fields are accessed only by the evaluator.

\endgroup

\section{Baseline Adaptation}
\label{app:adaptation}

Every comparison uses the same partition, timestamps, calibration, proposal pool, and evaluator. Adaptation supplies the 4DVLT input/output interface without removing native encoders, reducing backbone size, or replacing a baseline objective with the 4DTrack objective.

\paragraph{VLT trackers.}
Each synchronized clip is represented as per-camera sequences. JointNLT~\cite{zhouJointVisualGrounding2023}, UVLTrack~\cite{maUnifyingVisualVisionlanguage2024}, GLAD~\cite{zhangAwareDistillationRobust}, and DUTrack~\cite{liDynamicUpdatesLanguage} retain their language--appearance fusion and tracking objectives. A common multi-view adapter associates native 2D tracks with proposal states and passes the resulting 3D centers and visible-view support to the evaluator. Support is emitted only for successful track--proposal associations; the identical VLT coverage values therefore reflect this shared conservative association rule. Neither target identity nor the reference path is available at inference.

\paragraph{Text controls.}
Candidate features and edges are serialized in temporal order and mapped to the language-model hidden dimension. The controls include Llama-3-8B~\cite{grattafiori2024llama}, Mistral-7B~\cite{jiang2023mistral}, and Qwen3-8B~\cite{yang2025qwen3}; each is parameter-efficiently adapted to emit candidate-node sequences, which are parsed into the common worldline output. Accordingly, the open-source model rows in Table~\ref{tab:main_results} are task-adapted controls rather than zero-shot evaluations.

\paragraph{Multimodal controls.}
The synchronized visual input remains connected to each model's native vision encoder. This group includes Qwen2.5-VL-7B~\cite{qwen2.5-VL}, Qwen3.5-9B~\cite{qwen3.5}, and InternVL3.5-8B~\cite{wang2025internvl3}. A model-specific adapter combines its visual representation with the same candidate-state serialization used by the text controls, and the adapted decoder emits candidate-node sequences. The comparison therefore retains the pretrained modality pathway while standardizing only the task interface and output space.

\paragraph{Matched 4DTrack variants.}
For each backbone family, the matched 4DTrack configuration preserves the input evidence, parameter-efficient adaptation budget, sequential EgoWL-to-AlloWL schedule, and evaluator. It adds the graph projector, query-conditioned router, node decoder, direction-symmetric objective, and alignment head. Paired changes thus compare an adapted backbone control with structured worldline inference under the same task exposure.

\section{Computational Profile}
\label{app:profile_use}

Let $T$ be the retained-frame count, $K$ the candidate cap, $N\le TK$ the total nodes, $M$ the per-frame routing budget, and $B$ the beam size. Adjacent-frame graph construction costs $O(TK^2)$. The router scores $N$ nodes and passes at most $TM$ nodes to decoding. The reported setting uses $K=24$, $M=16$, and $B=4$. Its dominant costs are front-end proposal extraction, graph serialization, language-model decoding, and kinematic reranking. The present task is designed for offline calibrated analysis; streaming optimization~\cite{wang2026stc} is outside its scope.

Each table entry reports one fixed run. Experiments use eight NVIDIA A100 GPUs with 80\,GB memory per card, 128 vCPUs, and 1000\,GiB host memory under Ubuntu, CUDA 12.x, PyTorch 2.9, HuggingFace Transformers, PEFT, and bitsandbytes. The release package will include benchmark annotations and partitions, preprocessing and graph construction, training and inference, the evaluator, exact dependency versions, configuration files, and random seeds.

Instruct-4D is intended for research on calibrated scene understanding. Released assets remain subject to the source-dataset licenses and privacy constraints. The benchmark is not intended for identifying real individuals or for surveillance deployment.

\section{Additional Qualitative Cases}
\label{app:more_demos}

\begin{figure}[!htbp]
\centering
\includegraphics[width=0.96\linewidth]{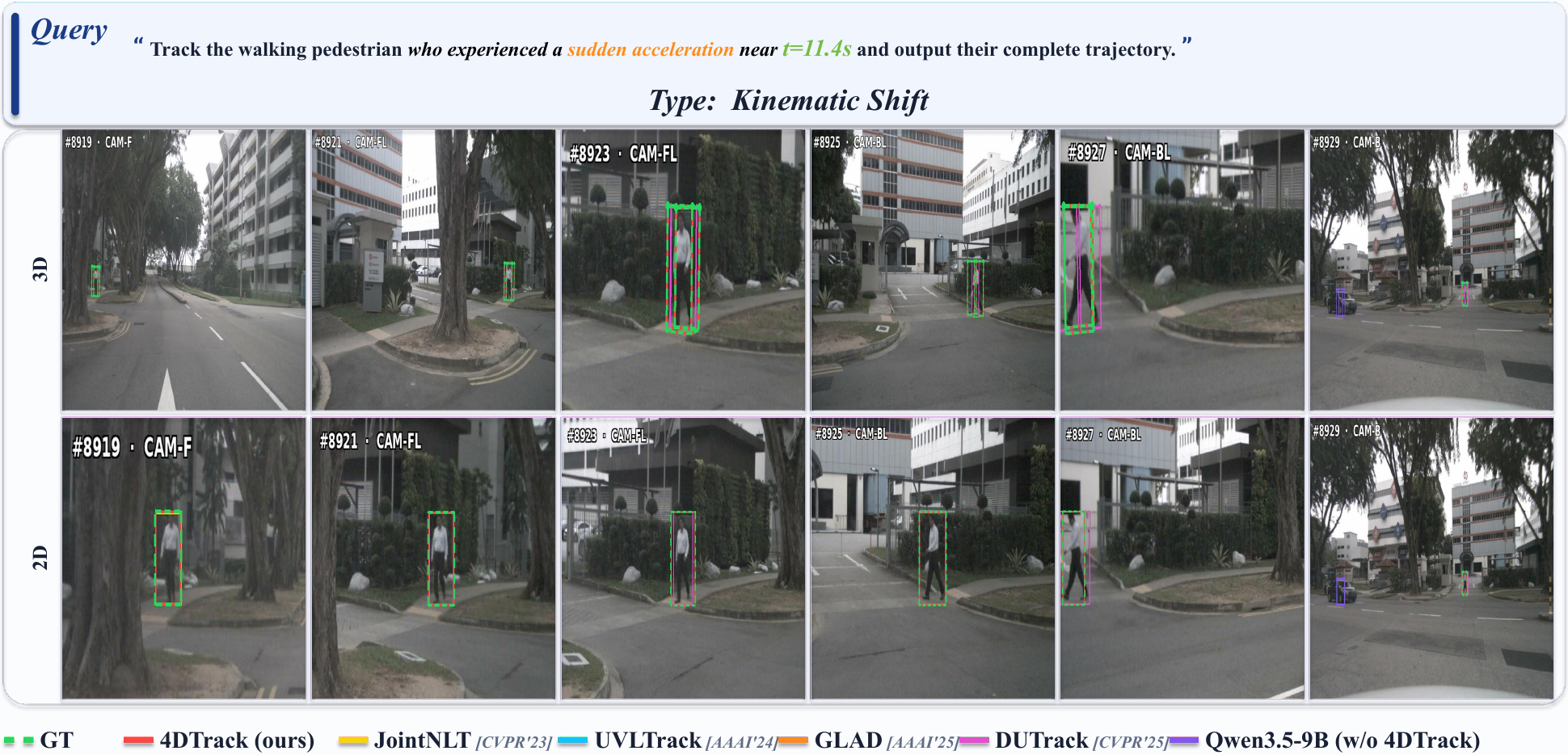}
\caption{Kinematic Shift (EgoWL). The referent is identified by a sudden acceleration near $t{=}11.4$\,s.}
\label{fig:demo_app_3}
\end{figure}

\begin{figure}[!htbp]
\centering
\includegraphics[width=0.96\linewidth]{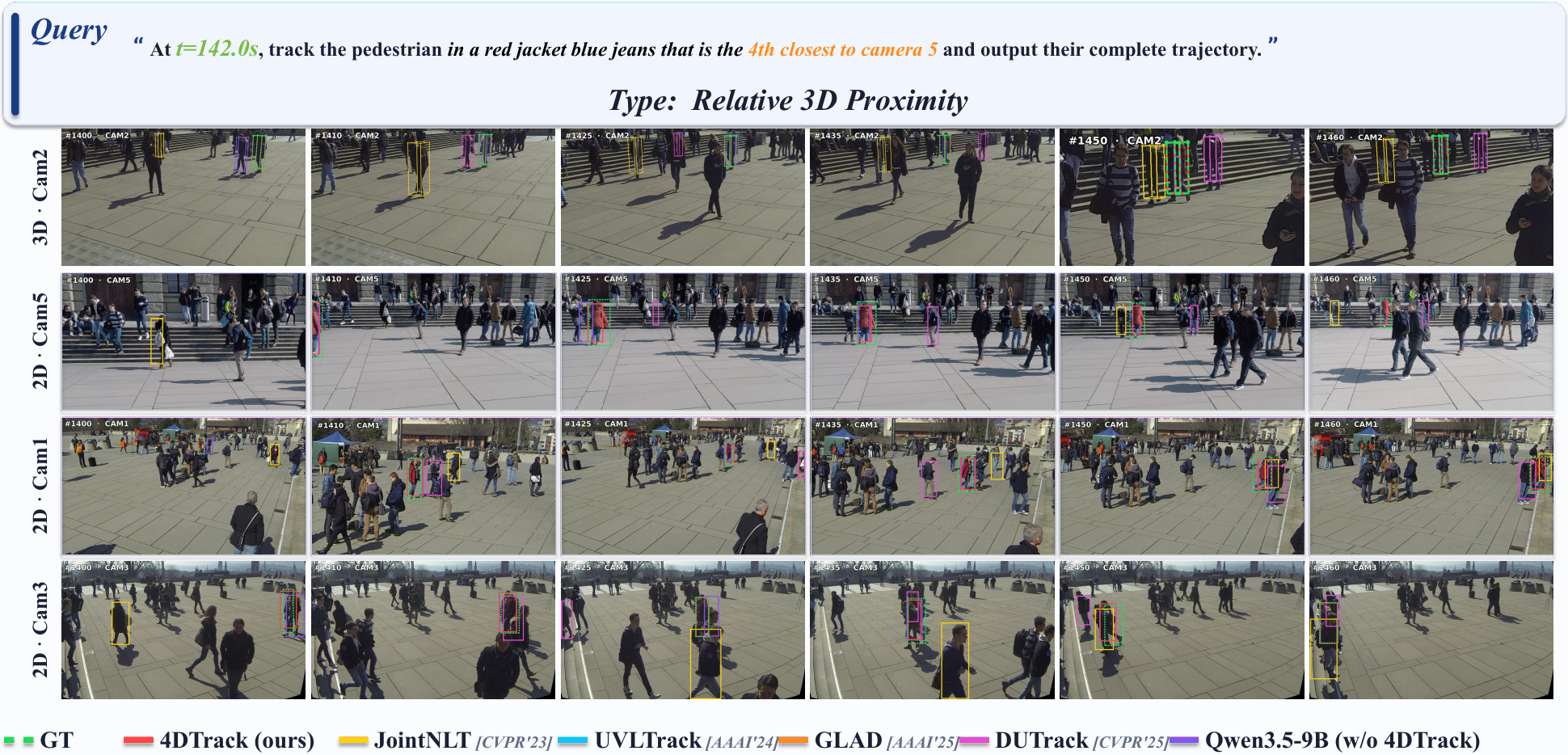}
\caption{Relative 3D Proximity (AlloWL). The target is the pedestrian ranked fourth closest to Camera 5 at $t{=}142.0$\,s.}
\label{fig:demo_app_4}
\end{figure}

\begin{figure}[!htbp]
\centering
\includegraphics[width=0.96\linewidth]{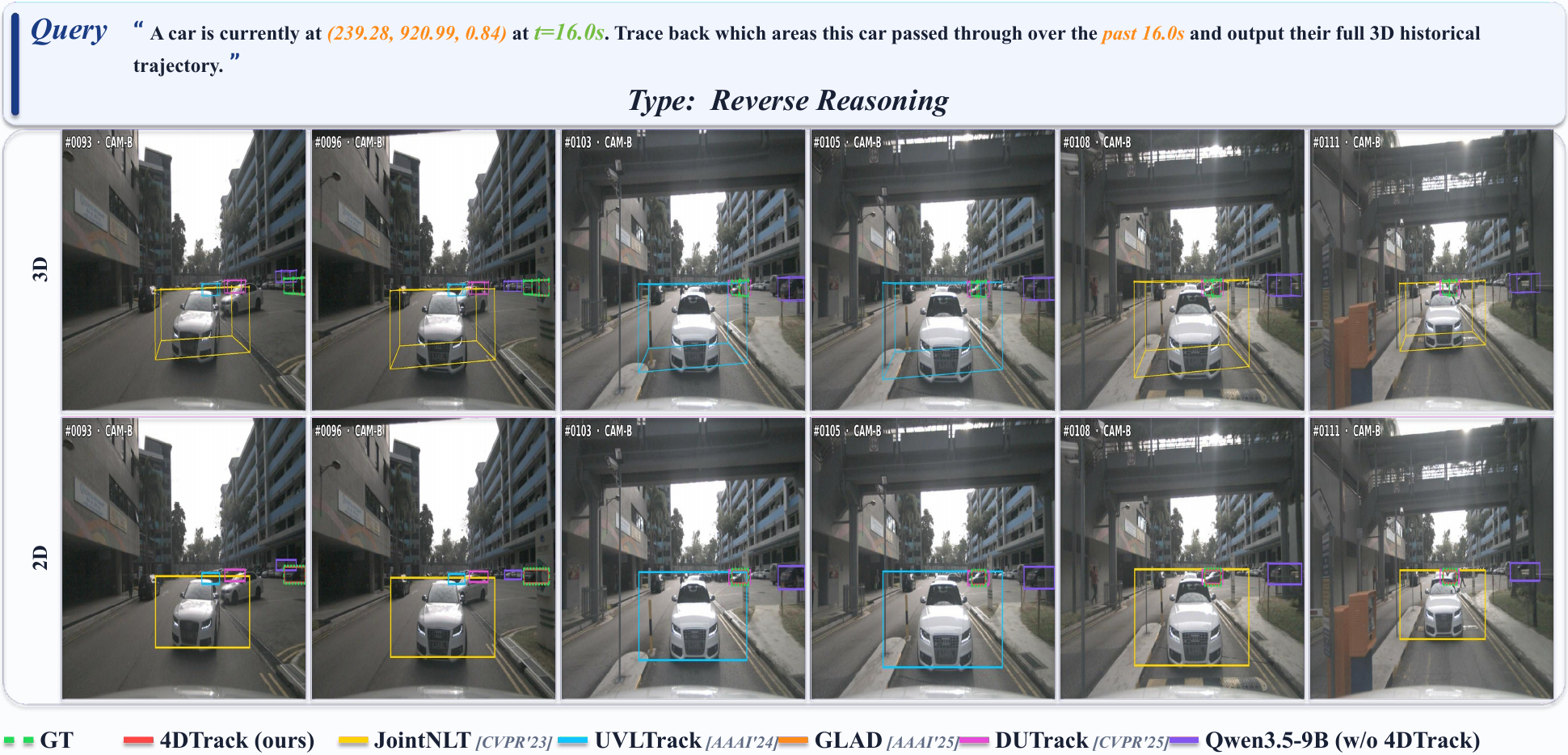}
\caption{Reverse Reasoning (EgoWL). A car observed at $t{=}16.0$\,s is traced through the preceding 16 seconds.}
\label{fig:demo_app_5}
\end{figure}

\begin{figure}[!htbp]
\centering
\includegraphics[width=0.96\linewidth]{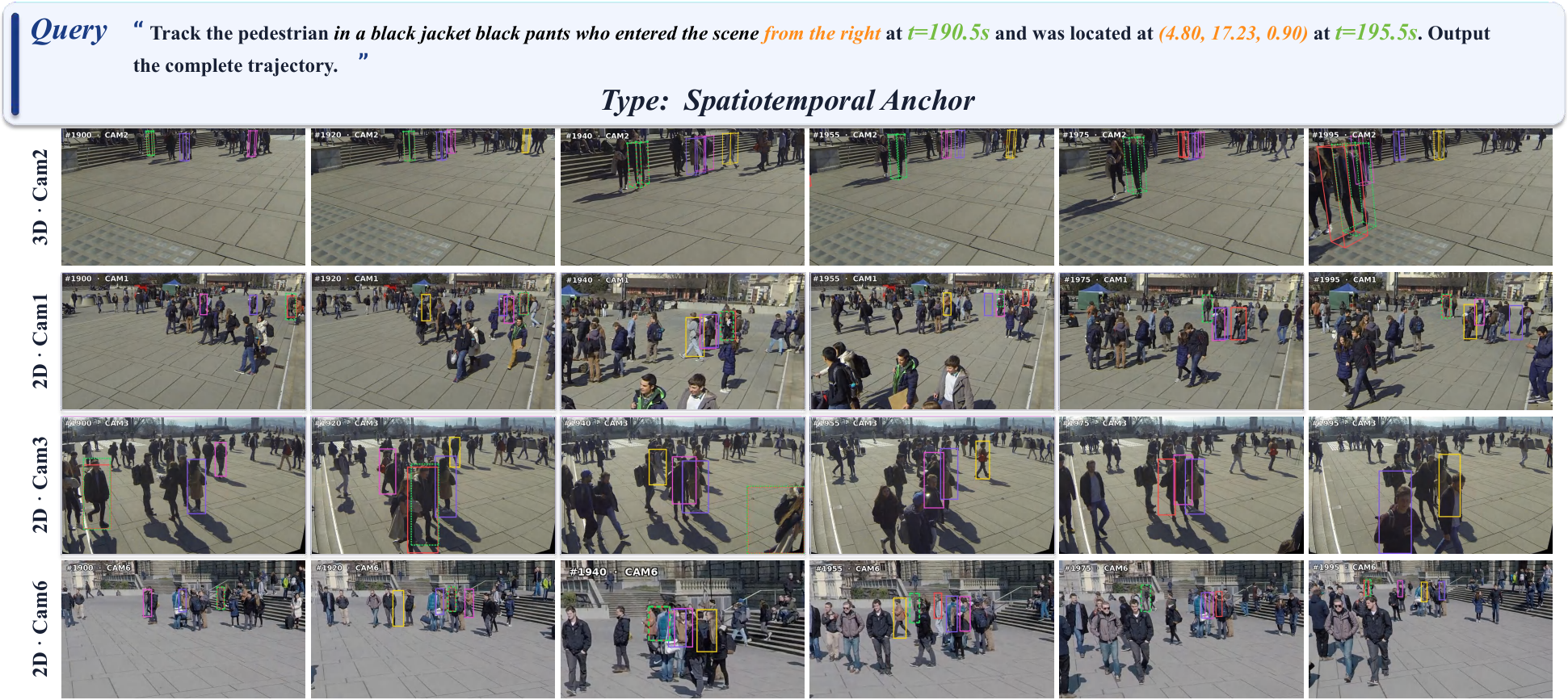}
\caption{Spatiotemporal Anchor (AlloWL). An entry event at $t{=}190.5$\,s is linked to a later metric position.}
\label{fig:demo_app_6}
\end{figure}

\begin{figure}[!htbp]
\centering
\includegraphics[width=0.96\linewidth]{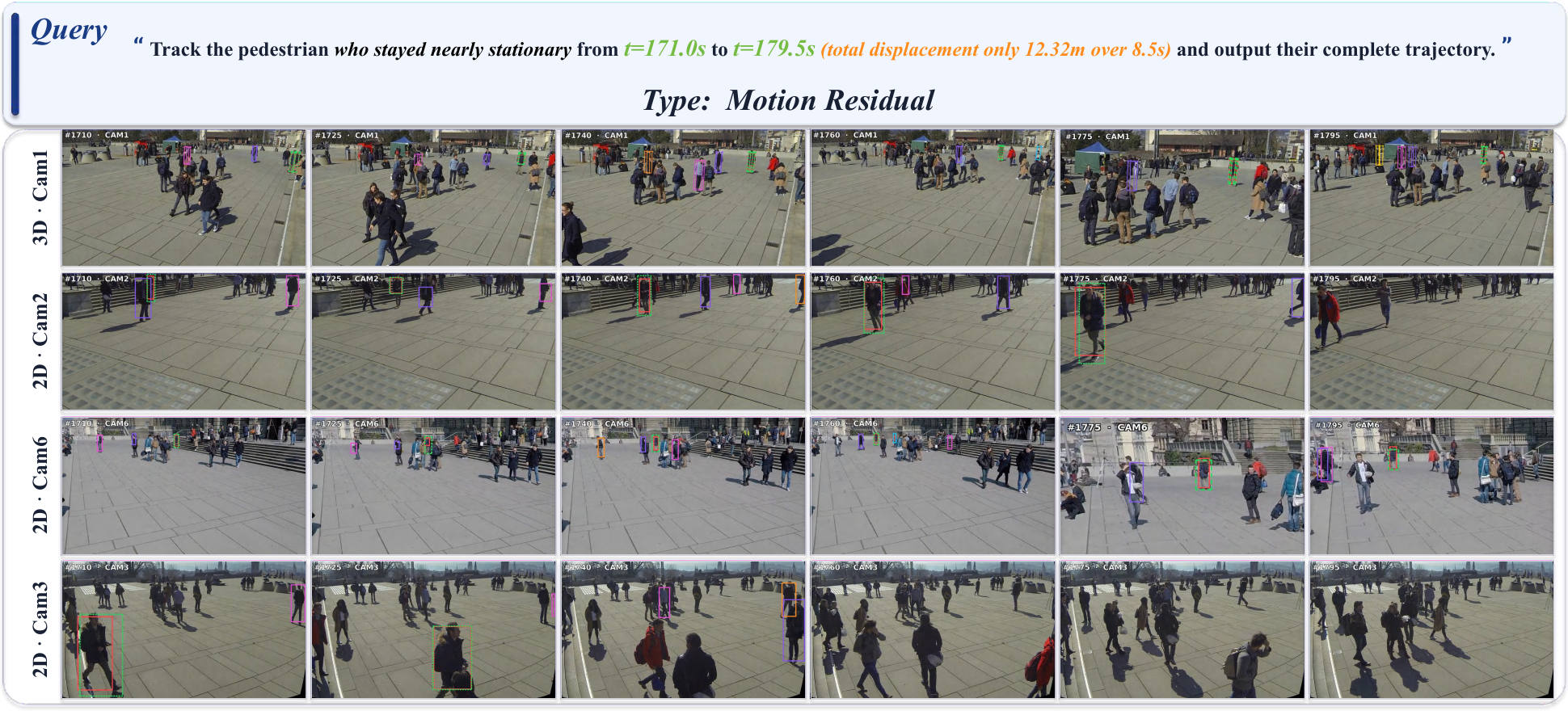}
\caption{Motion Residual (AlloWL). Residual displacement over $t{=}171.0$--$179.5$\,s identifies the target.}
\label{fig:demo_app_7}
\end{figure}

\begin{figure}[!htbp]
\centering
\includegraphics[width=0.96\linewidth]{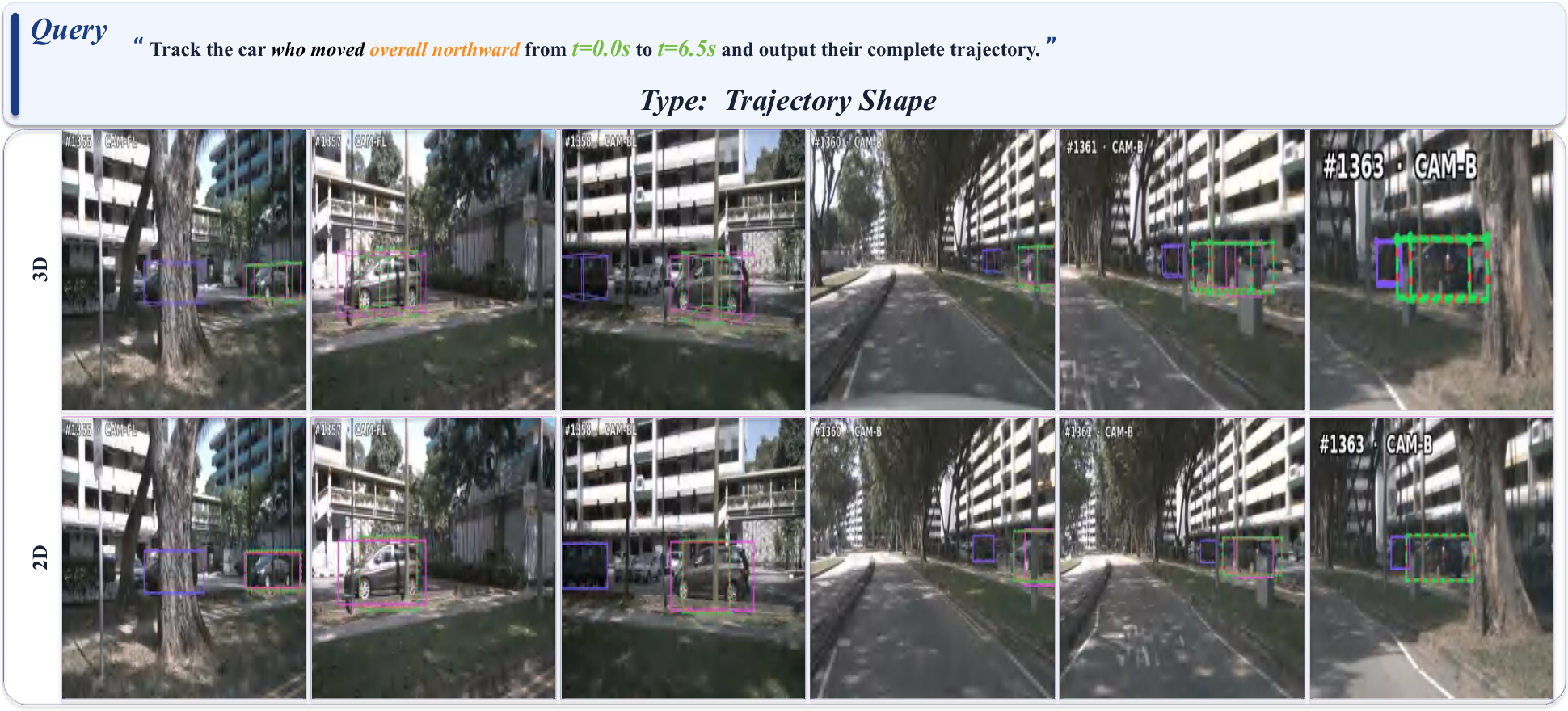}
\caption{Trajectory Shape (EgoWL). The instruction selects the car moving overall northward from $t{=}0.0$ to $6.5$\,s.}
\label{fig:demo_app_8}
\end{figure}

Figure~\ref{fig:demo} in the main paper shows 3D Volume Geometry. Figures~\ref{fig:demo_app_1}--\ref{fig:demo_app_8} cover the remaining query types with sampled 3D states and synchronized camera views.

\begingroup
\setlength{\intextsep}{4pt}
\setlength{\floatsep}{4pt}
\captionsetup{skip=4pt}
\begin{figure}[!htbp]
\centering
\includegraphics[width=0.96\linewidth]{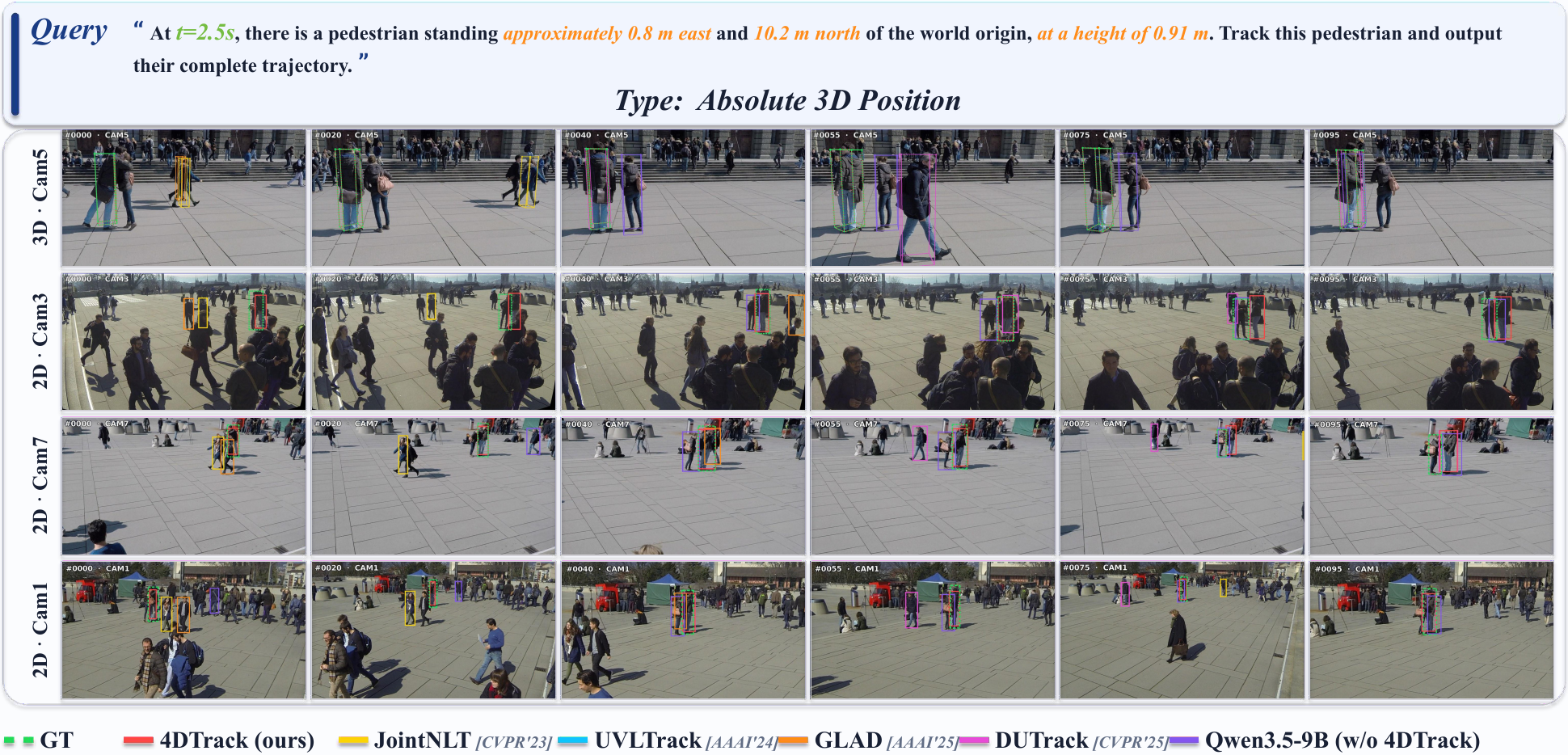}
\caption{Absolute 3D Position (AlloWL). A metric offset at $t{=}2.5$\,s identifies the pedestrian whose full worldline is requested.}
\label{fig:demo_app_1}
\end{figure}

\begin{figure}[!htbp]
\centering
\includegraphics[width=0.96\linewidth]{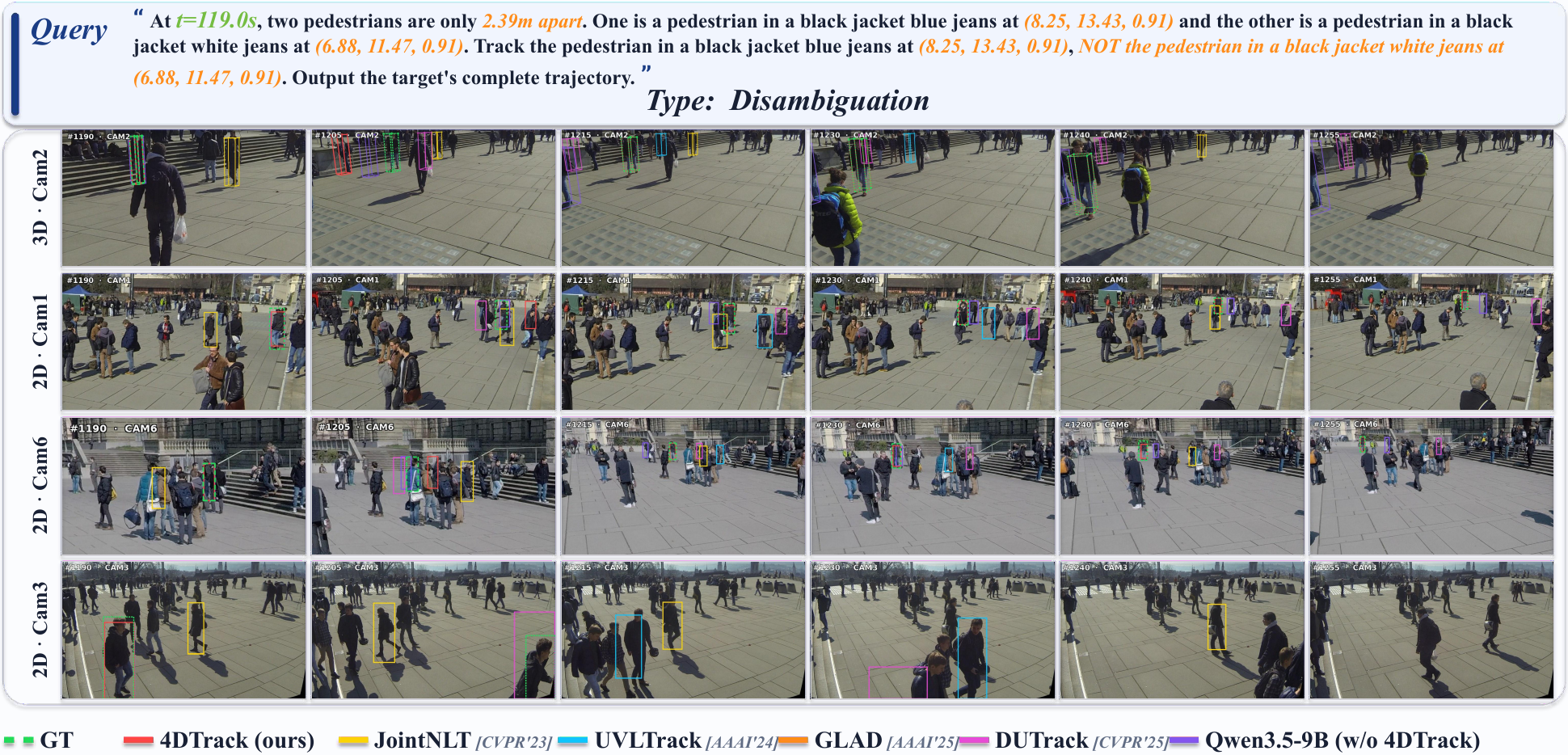}
\caption{Disambiguation (AlloWL). Appearance and position separate two similarly dressed pedestrians 2.39\,m apart.}
\label{fig:demo_app_2}
\end{figure}
\clearpage
\endgroup

\clearpage

\clearpage

\end{document}